\documentclass[compsoc, journal]{IEEEtran}
\usepackage{cite}
\usepackage{tabularx,booktabs}
\usepackage{amsmath,amssymb,amsfonts}
\usepackage{graphicx}
\usepackage{textcomp}
\usepackage[monochrome]{xcolor}
\usepackage{threeparttable}
\usepackage{placeins}
\usepackage{makecell}
\usepackage{enumitem}
\usepackage{soul}
\usepackage{verbatimbox}
\usepackage{csquotes}
\usepackage{multirow}
\usepackage{adjustbox}
\newcolumntype{s}{>{\hsize=.25\hsize \centering \arraybackslash}X}
\newcolumntype{Y}{>{\centering\arraybackslash}X}
\usepackage{float}
\usepackage{bbm}
\DeclareMathOperator*{\argmax}{argmax}
\DeclareMathOperator*{\argmin}{argmin}
\newtheorem{example}{Example}
\newtheorem{definition}{Definition}
\usepackage{algorithm}
\usepackage{algpseudocode}
\usepackage[framemethod=tikz]{mdframed}
\usepackage{hyperref}
\algnewcommand\algorithmicinput{\textbf{Input:}}
\algnewcommand\INPUT{\item[\algorithmicinput]}
\algnewcommand\algorithmicoutput{\textbf{Output:}}
\algnewcommand\OUTPUT{\item[\algorithmicoutput]}
\def\BibTeX{{\rm B\kern-.05em{\sc i\kern-.025em b}\kern-.08em
    T\kern-.1667em\lower.7ex\hbox{E}\kern-.125emX}}

\begin{document}
	
\setlength{\abovedisplayskip}{3.2mm}
\setlength{\belowdisplayskip}{3.2mm}

\title{Learning Entity Linking Features for \\ Emerging Entities}

% \author{\IEEEauthorblockN{Chenwei Ran}
% \IEEEauthorblockA{\textit{Tsinghua University \& Alibaba Group}\\
% Beijing, China \\
% ranchenwei@gmail.com}
% \and
% \IEEEauthorblockN{Wei Shen}
% \IEEEauthorblockA{\textit{Nankai University}\\
% Tianjin, China \\
% shenwei@nankai.edu.cn}
% \and
% \IEEEauthorblockN{Yuhan Li}
% \IEEEauthorblockA{\textit{~~~~~~~~~~~~~Nankai University~~~~~~~~~~~~~}\\
% Tianjin, China \\
% yuhanli@mail.nankai.edu.cn}
% \and
% \IEEEauthorblockN{Jianyong Wang}
% \IEEEauthorblockA{\textit{~~~~~~~~~~~~~~~~~~~~~~~~~~~~~Tsinghua University~~~~~~~~~~~~~~~~~~~~~~~~~~~~~}\\
% Beijing, China \\
% jianyong@tsinghua.edu.cn}
% \and
% \IEEEauthorblockN{Yantao Jia}
% \IEEEauthorblockA{\textit{Huawei Technologies Co., Ltd}\\
% Beijing, China \\
% jamaths.h@163.com}
% }

\author{
Chenwei Ran, Wei Shen, Jianbo Gao, Yuhan Li, Jianyong Wang, \IEEEmembership{Fellow,~IEEE}, Yantao Jia
\IEEEcompsocitemizethanks{
\IEEEcompsocthanksitem C. Ran and J. Wang are with the Department of Computer Science and Technology, Tsinghua University, Beijing 100084, China. J. Wang is also with the Jiangsu Collaborative Innovation Center for Language Ability, Jiangsu Normal University, Xuzhou, China. E-mail: ranchenwei@gmail.com, jianyong@tsinghua.edu.cn.
\IEEEcompsocthanksitem W. Shen, J. Gao and Y. Li are with TMCC, TKLNDST, the College of Computer Science, Nankai University, Tianjin 300350, China. E-mail: shenwei@nankai.edu.cn, \{jianbo.gao, yuhanli\}@mail.nankai.edu.cn.
\IEEEcompsocthanksitem Y. Jia is with the Huawei Technologies Co., Ltd, Beijing, China. E-mail: jamaths.h@163.com.
\IEEEcompsocthanksitem Wei Shen is the corresponding author.
}
}
\IEEEtitleabstractindextext{
\begin{abstract}
Entity linking (EL) is the process of linking entity mentions appearing in text with their corresponding entities in a knowledge base. EL features of entities (e.g., prior probability, relatedness score, and entity embedding) are usually estimated based on Wikipedia. However, for newly emerging entities (EEs) which have just been discovered in news, they may still \textbf{\textit{not}} be included in Wikipedia yet. As a consequence, it is unable to obtain required EL features for those EEs from Wikipedia and EL models will always fail to link ambiguous mentions with those EEs correctly as the absence of their EL features. To deal with this problem, in this paper we focus on a new task of learning EL features for emerging entities in a general way.
We propose a novel approach called \textbf{STAMO} to learn high-quality EL features for EEs automatically, which needs just a small number of labeled documents for each EE collected from the Web, as it could further leverage the knowledge hidden in the unlabeled data.
\textbf{STAMO} is mainly based on self-training, which makes it flexibly integrated with any EL feature or EL model, but also makes it easily suffer from the error reinforcement problem caused by the mislabeled data. Instead of some common self-training strategies that try to throw the mislabeled data away explicitly, we regard self-training as a multiple optimization process with respect to the EL features of EEs, and propose both intra-slot and inter-slot optimizations to alleviate the error reinforcement problem implicitly.
We construct two EL datasets involving selected EEs to evaluate the quality of obtained EL features for EEs, and the experimental results show that our approach significantly outperforms other baseline methods of learning EL features.
\end{abstract}

}
\maketitle

\section{Introduction}
\label{sec:intro}
Entity linking (EL), the task of mapping entity mentions in text to their corresponding entities in a target knowledge base (KB), is a fundamental step for various knowledge based applications \cite{shen2014entity,shen2021entity,sevgili2020neural, ma2019jointly, hao2017end, blanco2015fast}. In general, it is challenging mainly due to the universal many-to-many ambiguity between mentions and entities. On one hand, a named entity may have multiple names (e.g., full name, partial name, nickname, and abbreviation). For example, the named entity \textit{New York City} has a nickname \textit{Big Apple} and an abbreviation \textit{NYC}. On the other hand, a name could possibly denote different named entities. For instance, the name \textit{Michael Jordan} may refer to an NBA basketball player, a Berkeley machine learning professor, or many other named entities which could be referred to as \textit{Michael Jordan}. In recent years, EL models based on deep neural networks have achieved great success \cite{shen2021entity,sevgili2020neural}. For example, DeepED \cite{ganea2017deep} is a remarkable work which embeds entities and words in a common vector space and leverages a neural attention mechanism over local context windows, while other architectures such as graph neural networks \cite{cao2018neural, wu2020dynamic, fang2020high} and recurrent random walk networks \cite{xue2019neural} were investigated as well.

%shen revision
\begin{figure*}[thb]
%\begin{mdframed}[linecolor=blue,linewidth=1pt,innerrightmargin=1pt,innerbottommargin=-110pt,innerleftmargin=1pt,innertopmargin=0pt,backgroundcolor=white]
    \centering
    \includegraphics[width=0.98\textwidth]{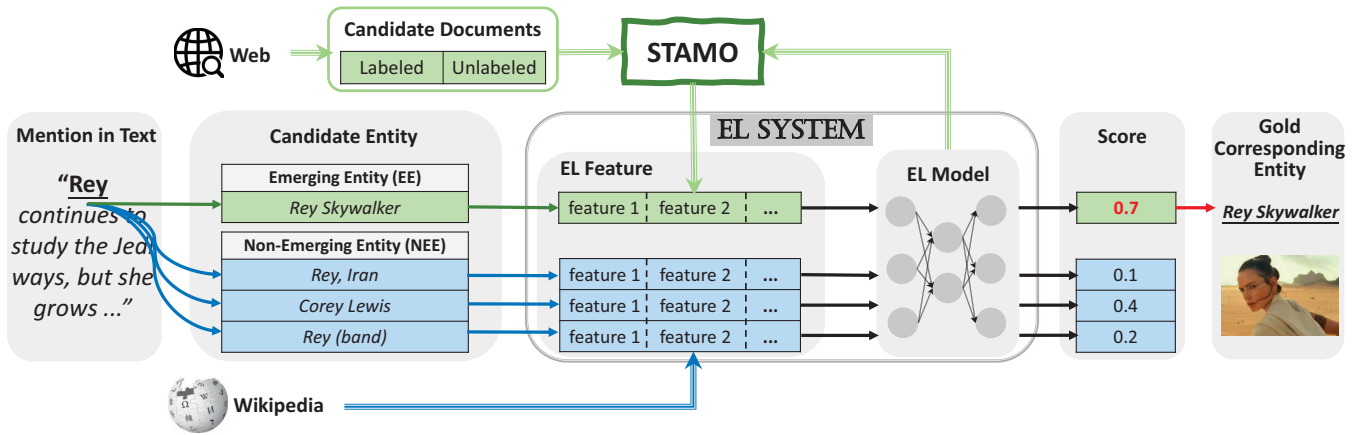}
    		\vspace{-3mm}
   \caption{An illustration for the task of learning entity linking features for an emerging entity. The ambiguous mention \textit{Rey} refers to four candidate entities, among which the EE \textit{Rey Skywalker} is the gold corresponding entity. EL features of the three non-emerging entities (NEEs) could be estimated from Wikipedia (highlighted in blue), whereas for the EE, we cannot obtain its EL features from Wikipedia. In this case, the EL model cannot calculate the score for this EE. To address this problem, the goal is to learn high-quality EL features for the EE based on the labeled and unlabeled candidate documents collected from the Web (highlighted in green). Then via leveraging the learned EL features for this EE, the EL model can calculate its score (i.e., 0.7), which is the highest among the candidates, and link the mention with this EE correctly.}
    \label{fig:illustration}
%    \end{mdframed}
\vspace{-5mm}
%    \vspace{-4mm}
\end{figure*}

Besides the choice of EL model, another key factor in EL systems is the way to obtain EL features of entities. \textcolor{blue}{As reviewed in \cite{shen2014entity,shen2021entity}, EL features of entities commonly utilized by EL models to rank candidate entities are prior probability \cite{ratinov2011local,hoffart2011robust,ganea2017deep,le2018improving,fang2020high, ran2018attention}, relatedness score \cite{hoffart2011robust, ratinov2011local, shen2012linden,phan2018pair, xue2019neural, hoffart2012kore}, and entity embedding \cite{shen2021entity,ganea2017deep,yamada2016joint,cao2017bridge,sevgili2019improving,fang2019joint,wu2020scalable,zwicklbauer2016robust}.
These EL features of entities} need to be estimated with the requirement of a labeled corpus where gold mention-entity pairs (i.e., pairs of mentions and their gold corresponding entities) are available \cite{shen2021entity,shen2014entity, ganea2017deep}. This labeled corpus should cover the entities in the KB as many as possible. If some entities never show up in the labeled corpus, it is unable to estimate required EL features for those entities. In practice, Wikipedia, a large-scale encyclopedia which contains billions of hyperlinks in its articles that could be regarded as gold mention-entity pairs for millions of entities, is leveraged by most EL works as the best choice of the labeled corpus for estimating EL features of entities \cite{shen2021entity,shen2014entity,sevgili2020neural}.

However, it is observed that almost 50\% of entities are first mentioned in news before they are included in Wikipedia, and the lags could be several months or even a year \cite{fetahu2015much}. That is to say, although these emerging entities (EEs) could be discovered from news and added to a KB with a unique identifier by hand or \textcolor{blue}{automatic methods like \cite{hoffart2014discovering}}, they cannot be linked with accurately by EL models because Wikipedia \textit{cannot} provide required EL features for these EEs.
To deal with this problem, in this paper we study a new task of learning EL features for emerging entities in a general way, which is illustrated in Figure \ref{fig:illustration}.

An intuitive idea to learn EL features for an EE is to collect some candidate documents where the EE may be mentioned from the Web, manually annotate them, and then estimate required EL features for the EE on these labeled data. However, this process is non-trivial in practice because manually annotating a large number of candidate documents for each EE is very time-consuming and labor-intensive. Additionally, a prior work \cite{singh2016discovering} proposed a retrieval-based approach to extract keyphrase descriptions for EEs. However, keyphrase description is not a strong EL feature, and their approach cannot be extended to other commonly used EL features (e.g., prior probability, relatedness score, and entity embedding). It is also noted that zero-shot learning \cite{romera2015embarrassingly} has already been considered in the field of entity linking. In the setting of zero-shot EL \cite{li2020efficient, logeswaran2019zero, wu2020scalable}, the entity being linked with has not been seen during training and is defined by a textual entity description from its corresponding Wikipedia page. Zero-shot EL models rely on language understanding of the given textual entity description to perform entity linking. However, this setting makes these zero-shot EL models unable to deal with our problem in this paper, as the EE has not been included in Wikipedia yet and does not have the textual entity description accordingly.

Overall, our goal is to learn high-quality EL features for the EE automatically.
The proposed approach is expected to be applicable for any numerical EL feature (e.g., prior probability, relatedness score, and entity embedding) rather than a specific one, as long as the EL feature could be automatically estimated given a labeled corpus.
It is also noted that this new task is defined as an orthogonal effort to the task of developing an EL model that most previous EL works focus on.
This makes the proposed approach flexibly integrated with any state-of-the-art EL models such as DeepED \cite{ganea2017deep} \textcolor{blue}{and Yamada et al. \cite{yamada2016joint}}, while the prior work \cite{singh2016discovering} is restricted to keyphrase-based EL models \cite{hoffart2011robust} only.

In this paper, we propose a novel approach called \textbf{STAMO} (\textbf{S}elf-\textbf{T}raining \textbf{A}s \textbf{M}ultiple \textbf{O}ptimizations) satisfying all the above requirements. \textbf{STAMO} is mainly based on self-training, a classic branch of semi-supervised learning. In general case, self-training iteratively assigns pseudo labels to the unlabeled data by a classification model and updates this classification model on both real labeled and pseudo labeled data. While in our case, the pseudo label means the entity that a mention refers to, the classification model is an EL model, and \textbf{the target to update} is the EL features of EEs (rather than the EL model) based on both labeled and unlabeled candidate documents collected from the Web, as shown in Figure \ref{fig:illustration}. \textbf{STAMO} needs just a small number of labeled documents for each EE, as it could further leverage the knowledge hidden in the unlabeled data. In addition, \textbf{STAMO} learns EL features for EEs \textit{incrementally}. That is to say, given some \textcolor{blue}{non-emerging entities (NEEs)} whose EL features have been estimated from Wikipedia and a trained EL model, we do not need to update the EL features of NEEs and the EL model every time when an EE is discovered.

\vspace{1mm}
The main advantage of self-training is that it is a wrapper method, which makes \textbf{STAMO} flexibly integrated with any numerical EL feature or EL model. On the other hand, the major drawback of self-training is that the errors introduced by the mislabeled data may be reinforced gradually \cite{zhu2009introduction}. In our task, we found that the vanilla self-training method collapses quickly after a few slots\footnote{In this paper, we will also call an iteration in self-training as a \textit{slot}.}, i.e., the quality of EEs' EL features becomes worse after update. To alleviate this problem, a common assumption is that high confident predictions tend to be correct \cite{triguero2015self}, and only a few unlabeled instances with the most confident predictions will be combined with the labeled data in the iterative process of self-training. However, the prediction score output by some EL models (e.g., DeepED \cite{ganea2017deep} and Yamada et al. \cite{yamada2016joint}) does not act as a competent measure of confidence, which is possibly caused by the use of max-margin loss. At the same time, the selection of instances may introduce additional bias, which would significantly affect the quality of some statistics-based features (e.g., prior probability). Therefore, common strategies based on this assumption are unsuitable for our task, and we keep all unlabeled data in all slots as suggested in \cite{zhu2009introduction} to update the EL features.

In \textbf{STAMO}, we creatively regard self-training as a multiple optimization process with respect to the EL features of EEs, which could alleviate the error reinforcement problem caused by the mislabeled data implicitly, rather than throw the mislabeled data away explicitly. Specifically, we propose both intra-slot optimization and inter-slot optimization. In intra-slot optimization, we regard the EL features of EEs estimated over the mix of the real labeled and pseudo labeled data as an initial guess instead of the final result, and consider that the EL features of EEs should also minimize the objective function which is used to train the given EL model only based on the real labeled data. In inter-slot optimization, we propose a hypothetical optimization process which makes a direct connection between the EL features of EEs in the current slot and the historical slots, and this novel perspective enables us to leverage the information provided by the historical slots to improve the future learning process.

\textbf{Contributions.} The main contributions of this paper can be summarized as follows:

\begin{itemize}
    \item To the best of our knowledge, we are the first to study the task of learning EL features for EEs in a general way, an orthogonal effort to the task of developing an EL model that most previous EL works focus on.
    \item We propose a novel approach \textbf{STAMO} based on self-training, which makes it flexibly integrated with any numerical EL feature or EL model. We regard self-training as a multiple optimization process, and propose both intra-slot and inter-slot optimizations to alleviate the error reinforcement problem implicitly.
    \item In inter-slot optimization, we propose a novel hypothetical optimization process, which enables us to leverage the information provided by the historical slots to improve the future learning process.
    \item We construct two EL datasets to evaluate the quality of the obtained EL features for EEs, and the experimental results show that our approach significantly outperforms other baseline methods of learning EL features.
\end{itemize}

%%%%%%%%%%%%%%%%%%%%%%%%%%%%%%%%%%%%%%%%
\vspace{-2mm}
\section{Preliminaries}

%shen revision
\subsection{Notations and Problem Definition}
\label{sec:Preliminaries}

% We begin with some basic concepts in EL. Given a document $d$, we assume a set of $K$ mentions $\mathbf{m} = \{m_1, m_2, ..., m_K\}$ has been detected (e.g., \textit{Rey} in Fig. \ref{fig:illustration}). For a mention $m_i$, a fixed length sequence of words surrounding it is collected as its context $c_i$. The goal of EL is mapping these mentions to their corresponding entities (e.g., \textit{Rey Skywalker} in Fig. \ref{fig:illustration}) in a target KB which comprises a set of entities $E$. Each entity $e \in E$ could have different aliases, which set is denoted by $A(e)$. According to the alias dictionary, we can identify a set of candidate entities $E(m) = \{e: m \in A(e)\}$ (e.g., \textit{Rey Skywalker}, \textit{Rey (band)}, etc. in Fig. \ref{fig:illustration}) for a mention $m$. In real-world data, it is usually found that $\overline{|A(e)|} > 1$ and $\overline{|E(m)|} > 1$ simultaneously, which inevitably brings the many-to-many ambiguity problem.
We begin with some basic concepts in EL. Given a document $d$, we assume a set of $K$ mentions $\mathbf{m} = \{m_1, m_2, ..., m_K\}$ has been detected (e.g., \textit{Rey} in Figure \ref{fig:illustration}) via named entity recognition methods \cite{li2021sequence, li2020few, li2020survey}. The goal of EL is mapping these mentions to their corresponding entities (e.g., \textit{Rey Skywalker} in Figure \ref{fig:illustration}) in a target KB which comprises a set of entities $E$. Each entity $e \in E$ could have different aliases, and the set of its aliases is denoted by $A(e)$. According to the alias dictionary, we can identify a set of candidate entities $E(m) = \{e: m \in A(e)\}$ for a mention $m$ (e.g., \textit{Rey Skywalker} and \textit{Rey (band)} for the mention \textit{Rey} in the aforementioned example). In real-world data, it is usually found that $|A(e)| > 1$ and $|E(m)| > 1$ simultaneously, which inevitably brings the many-to-many ambiguity problem.

\begin{table}[t]
 \centering
 \caption{Notations}
 		\vspace{-2mm}
 \begin{tabular}{|c|c|}
  \hline
  Symbol&  Description \\
  \hline
   \hline
  $d$ & A document \\
  \hline
  $m$ & A mention in $d$ \\
  \hline
  $E$ & A set of entities in a KB\\
  \hline
  $e$&  An entity in $E$\\
  \hline
  $\hat{e}$  &  The predicted corresponding entity for mention $m$\\
  \hline
  $e^\dag $  &  The gold corresponding entity for mention $m$\\
  \hline
  $e^*$&  An emerging entity (EE) \\
  \hline
  $\overline{e^*}$&  A non-emerging entity (NEE)\\
  \hline
  $E(m)$ & A set of candidates entities for mention $m$\\
  \hline
  $A(e)$ &  A set of aliases for entity $e$\\
  \hline
  $g$ & An  EL model\\
  \hline
  $\phi$ & EL model parameters w.r.t. EL model $g$\\
  \hline
  $\hat{\phi}$ & Optimal EL model parameters w.r.t. EL model $g$\\
  \hline
  $f$ &  EL feature function\\
  \hline
  $\theta$ & EL feature parameters\\
  \hline
  $\hat{\theta}$ & Optimal EL feature parameters\\
  \hline
  $\theta(e)$ & EL feature parameters associated with entity $e$\\
  \hline
  $\theta^t$ & EL feature parameters at iteration $t$\\
  \hline
  $W$ & A set of candidate documents\\
  \hline
  $L$ & A labeled subset of $W$\\
  \hline
  $U$ & An unlabeled subset of $W$\\
  \hline
  $\mathcal{O}$  & Objective function w.r.t. EL model $g$ \\
  \hline
  $c$ & A word context surrounding mention $m$\\
  \hline
  $\mathbf{\hat{e}}$  & An entity context surrounding mention $m$\\
  \hline
  \end{tabular}
 \label{tab:notation}
 \vspace{-4mm}
\end{table}

To identify the corresponding entity for mention $m_i$, a real-number probability or score is assigned to each candidate entity $e_i \in E(m_i)$ by an EL model $g$ parameterized by $\phi$ and the candidate entity with the highest probability or score is output as the linking result. In general, EL model can be divided into two categories: local model and global model. A local model $g:E(m_i) \rightarrow R$ links one mention at a time. It predicts the corresponding entity for mention $m_i$ by the following formula:
$$
	\hat{e}_i = \argmax_{e_i \in E(m_i)}  g_\phi(e_i ; m_i)
$$
\noindent where $\hat{e}_i$ denotes the predicted corresponding entity for mention $m_i$. Therefore, we need to repeat it for $K$ mentions in document $d$ such that every mention is linked. A global model $g:E(m_1) \times ... \times E(m_K) \rightarrow R$ collectively links all mentions in document $d$ at the same time by the following formula:
$$
	\hat{e}_1, ..., \hat{e}_K = \argmax_{e_1 \in E(m_1), ..., e_K \in E(m_K)}  g_\phi(e_1, ..., e_K ; \mathbf{m})
$$
Given a labeled corpus $S$ with gold mention-entity pairs $\{(m, e^\dag)\}$, model parameters $\phi$ for the EL model $g$ can be optimized with respect to a suitable objective function $\mathcal{O}$. The optimal model parameters are denoted by $\hat{\phi}$.
% Since we want to learn EL features for EEs incrementally, model parameters $\phi$ will be frozen all the time in our proposed approach.

Next, we introduce more details about EL features for entities. In the context of EL, an EL feature for entities means a table-lookup feature function $f:E \times ... \times E \rightarrow \mathbb{R}$ parameterized by $\theta$. In this paper, we call $\theta$ feature parameters in order to distinguish them from general model parameters $\phi$ introduced above. Here we give two examples as follows:

\begin{example}
	The feature function of entity embedding $f:E \rightarrow \mathbb{R}^Q$ maps an input entity to a $Q$-dimensional real vector space. It can be defined as $f(e) = \theta \mathbf{x}_{e}$, where $\mathbf{x}_{e} \in \mathbb{R}^{|E|}$ is one-hot encoding of entity $e$ and $\theta \in \mathbb{R}^{Q \times |E|}$ is the corresponding feature parameters.
\end{example}

\begin{example}
	The feature function of relatedness score $f:E \times E \rightarrow R$ maps two entities jointly to a real number. It can be defined as $f(e, e') = \mathbf{x}_{e}^\intercal \theta \mathbf{x}_{e'}$, where $\theta \in \mathbb{R}^{|E| \times |E|}$ is the corresponding feature parameters.
\end{example}

\noindent In practice, usually more than one feature would be leveraged. The corresponding feature parameters are grouped together as $\theta = (\theta_1, \theta_2, ...)$. The part of $\theta$ that associates with an entity $e$ is denoted by $\theta(e)$.
% Before training the EL model, the best feature parameters for NEEs were estimated on Wikipedia.

Then, an EE which has just been discovered and added to a KB is denoted by $e^*$. $e^*$ has not been included in Wikipedia, and the only information known about EE $e^*$ is its aliases $A(e^*)$. According to aliases $A(e^*)$, the mentions that may refer to EE $e^*$ could be identified. These mentions are called \emph{candidate mentions}, while other mentions that cannot refer to EE $e^*$ are called \emph{non-candidate mentions}. A set of candidate documents $W$ each of which contains at least one candidate mention is collected from the Web, and a small subset $L \subseteq W$ is manually annotated (i.e., the gold corresponding entities for candidate mentions in $L$ have been labeled manually). Here, it is noted that the gold corresponding entity of a candidate mention is not always the EE, but may be another entity that has the same mention as this EE, as the candidate mention may be ambiguous. The remaining unlabeled data is denoted by $U$, whose size is usually much larger than the labeled data (i.e., $|U| >> |L|$). Besides candidate mentions, we assume all the non-candidate mentions in both labeled data $L$ and unlabeled data $U$ have been linked to their corresponding entities by an existing EL system \textcolor{blue}{(e.g., DeepED \cite{ganea2017deep} and Yamada el al. \cite{yamada2016joint}). In the following, we define our task of learning EL features for EE  formally.}

\begin{definition}[\textcolor{blue}{\textbf{EL feature learning for EE}}]
\textcolor{blue}{Given an emerging entity $e^*$, a set of candidate documents $W$ of which a subset $L$ is manually labeled and an EL model $g$, the goal of our task is to learn the optimal EL feature parameters $\hat{\theta}(e^*)$ for EE $e^*$.}
\end{definition}

Note that since we want incremental learning, only feature parameters for EE $\theta(e^*)$ is modifiable and trainable in our proposed approach, while the optimal model parameters $\hat{\phi}$ has been learned with non-emerging entities (NEEs) only, and the optimal feature parameters for NEEs $\hat{\theta}(\overline{e^*})$ have been estimated on Wikipedia, both of which will be frozen all the time. Some notations used in this paper are summarized in Table \ref{tab:notation}.

\subsection{Entity Linking Feature}
\label{sec:entitylinkingfeature}

In this section, we introduce three widely used EL features: prior probability, relatedness score, and entity embedding. The estimation methods and incremental update rules for the corresponding feature parameters are stated as follows.

\vspace{0.2cm}
\noindent \textbf{(1) Prior Probability}

Prior popularity is the probability of the appearance of a candidate entity given an entity mention without considering the context where the mention appears. It has been adopted by most EL methods \cite{shen2021entity,ratinov2011local,hoffart2011robust,ganea2017deep,le2018improving,fang2020high} since it is a simple but strong signal in EL. In many real-world datasets, the accuracy of using prior probability alone can reach more than 80\% \cite{ganea2017deep}. Specifically, its parameters $\theta_1 \in \mathbb{R}^{|E| \times |M|}$ are estimated on Wikipedia article corpus (denoted by $S_{wiki}$) as follows:
\begin{equation}
	\theta_1^{i,j} = P(e^i|m^j) = \frac{\mbox{count}(e^i, m^j; S_{wiki})}{\sum_{e \in E(m^j)}\mbox{count}(e, m^j; S_{wiki})}
\end{equation}
\noindent where $\mbox{count}(e, m; S_{wiki})$ denotes the total times of entity $e$ referred by mention $m$ in the Wikipedia corpus $S_{wiki}$, $e^i$ and $m^j$ denote the $i$-th entity in $E$ and $j$-th mention in $M$ (the set of all mentions) respectively, and $\theta^{i,j}_1$ denotes the element of $\theta_1$ with index $(i, j)$.

For EE $e^*$, without loss of generality, we assume it is the last entity in $E$ (i.e., $e^* = e^{|E|}$). Given some labeled candidate documents (they could be either the real labeled data or the mix of the real labeled and pseudo labeled data) for $e^{|E|}$ collected from the Web, we denote them by $S_{web}$ and the parameters associated with $e^{|E|}$ are incrementally updated as follows:
\begin{equation}
	\theta_1^{|E|, j} = \frac{\mbox{count}(e^{|E|}, m^j; S_{web})}{\sum_{e \in E(m^j)}\mbox{count}(e, m^j; S_{web})}
\end{equation}
\noindent Finally, the corresponding feature function is defined as $f_1(e, m) = \mathbf{x}^\intercal_e \theta_1 \mathbf{x}_m$, where $\mathbf{x}_{e} \in \mathbb{R}^{|E|}$ is one-hot encoding of entity $e$, and $\mathbf{x}_m \in \mathbb{R}^{|M|}$ is one-hot encoding of mention $m$.

\vspace{0.2cm}
\noindent \textbf{(2) Relatedness Score}

Relatedness score is usually used to calculate the topical coherence between candidate entities in EL. Wikipedia Link-based Measure (WLM) \cite{witten2008effective} is the most widely utilized relatedness score in existing EL systems \cite{hoffart2011robust, ratinov2011local, shen2012linden,phan2018pair, xue2019neural}. It is defined under the assumption that two entities tend to be related if there are many documents that link to both. Here, we choose WLM as the relatedness score between entities and define it as follows:
\begin{equation}
	\begin{split}
	WLM&(e^i, e^j; S) = \\
	& 1 - \frac{log( max( |D(e^i)|, |D(e^j)| ) - log( |D(e^i)\cap D(e^j)| ) )}{log(|D|) - log( min( |D(e^i)|, |D(e^j)| ) )}	
	\end{split}
\label{eq:rsdefine}
\end{equation}
\noindent where $D$ denotes the set of all documents in the corpus $S$ and $D(e)$ denotes the set of documents where entity $e$ is referred. Therefore, its parameters $\theta_2 \in \mathbb{R}^{|E| \times |E|}$ are estimated on Wikipedia as $\theta_2^{i,j} = WLM(e^i, e^j; S_{wiki})$. For the EE $e^* = e^{|E|}$, we incrementally update the parameters in the $|E|$-th row as $\theta_2^{|E|,j} = WLM(e^{|E|}, e^j; S_{web})$ and the parameters in the $|E|$-th column as $\theta_2^{j, |E|} = WLM(e^{j}, e^{|E|}; S_{web})$. The feature function is defined as $f_2(e, e') = \mathbf{x}_{e}^\intercal \theta_2 \mathbf{x}_{e'}$.
%$D(i)$ denotes the set of documents where entity $e^i$ is referred in Wikipedia. $\theta_2$ is a symmetric matrix, i.e., $\theta_2^{i,j} = \theta_2^{j,i}$. We incrementally update the parameters associated with $e^{|E|}$ by setting the $|E|$-th row and $|E|$-th column of $\theta_2$ according to Eq. \ref{eq:rsdefine}. The feature function is defined as $f_2(e, e') = \mathbf{x}_{e}^\intercal \theta_2 \mathbf{x}_{e'}$.

\vspace{0.2cm}
\noindent \textbf{(3) Entity Embedding}

Entity embedding that represents entities in a continuous vector space has been developed rapidly in recent years and leveraged by many deep learning based EL models \cite{shen2021entity,ganea2017deep,yamada2016joint,cao2017bridge, sevgili2019improving,fang2019joint,wu2020scalable,zwicklbauer2016robust} to calculate the context similarity and the topical coherence between candidate entities. For this feature, the $i$-th column of its parameters $\theta_3 \in \mathbb{R}^{Q \times |E|}$ is set to the entity embedding of $e^i$, and the feature function is defined as $f_3(e) = \theta_3 \mathbf{x}_e$. We estimate entity embeddings according to the method proposed in DeepED \cite{ganea2017deep}. \textcolor{blue}{Specifically, for an entity, we first collect its co-occurrence words from the Wikipedia corpus $S_{wiki}$ and regard them as positive words of that entity. In addition, we randomly sample negative words unrelated to that entity in $S_{wiki}$. Finally, we use a max-margin loss to infer the optimal embedding of the entity with a goal that embeddings of positive words are closer to the embedding of that entity compared with embeddings of negative words.} For the EE $e^* = e^{|E|}$, we incrementally estimate its embedding based on $S_{web}$.

% We estimate entity embedding according to the method proposed in \cite{ganea2017deep}. The $i$-th column of its parameters $\theta_3 \in \mathbb{R}^{Q \times |E|}$ is set to the entity embedding of $e^i$, and the feature function is defined as $f_3(e) = \theta_3 \mathbf{x}_e$. Entity embedding supports incremental update instinctively. In addition, for the completeness of content, we define a feature function $f_4(w):V \rightarrow \mathbb{R}^Q$ that maps a word $w \in V$ to a $Q$-dimensional embedding. Word embeddings are used in both estimating entity embeddings and computing context similarity scores which will be described in the next subsection. We set $Q=300$ and word2vec pretrained embeddings \cite{mikolov2013distributed} were used in our experiments.
\subsection{Entity Linking Model}
\label{section:model}

In our experiments, we choose two mainstream and advanced EL models to work with our \textbf{STAMO}. One is the EL model proposed in Yamada et al. \cite{yamada2016joint}, and the other is an extended variation of the local EL model proposed in DeepED \cite{ganea2017deep}. In this section, we give a brief introduction of these two selected EL models and the learning strategy.

\subsubsection{\textcolor{blue}{Yamada}}
\label{section:yamada}

\textcolor{blue}{The model proposed by Yamada et al. \cite{yamada2016joint} (called Yamada in the remainder) is an influential EL model, which mainly relies on several similarity scores to perform linking.}

\textcolor{blue}{We formally define the word context $c$ of a candidate mention $m$ that needs to be linked as $c = \{w_1,...,w_H\}$. Since other non-candidate mentions in the same document can be linked by existing EL systems, we consider their corresponding entities as an entity context $\mathbf{\hat{e}} =\{\hat{e}_1,...,\hat{e}_K\}$ for the mention $m$. Based on the above definitions, we introduce six kinds of scores leveraged by Yamada in the following. The first one is the prior probability $\Psi^{(Y)}_{1}(e, m) = f_1(e, m)$ introduced in Section \ref{sec:entitylinkingfeature}, which captures the probability of the appearance of an entity given a mention without considering the context. The second score is the context similarity, which captures the similarity between the entity and the words in the context. Following the same setting in Yamada, word embeddings are trained to align with entity embeddings, and a word in the vocabulary (denoted by $w \in V$) is mapped to its embedding by a feature function $f^{(Y)}_4(w):V \rightarrow \mathbb{R}^Q$. Hence, the context similarity score $\Psi^{(Y)}_{2}(e, c)$ is derived as follows: }

\begin{equation}
    \textcolor{blue}{\Psi^{(Y)}_{2}(e, c) = cos(f_3(e), \frac{1}{H} \sum_{w \in c}^{} f^{(Y)}_4(w))}
\end{equation}

\noindent \textcolor{blue}{where $cos(\cdot)$ denotes the cosine similarity measure and $H$ is the number of words in the word context $c$. The third score is the topical coherence, which captures the similarity between the candidate entity and the entity context based on entity embeddings. The topical coherence score $\Psi^{(Y)}_{3}(e, \mathbf{\hat{e}})$ is defined as follows:}

\begin{equation}
    \textcolor{blue}{\Psi^{(Y)}_{3}(e, \mathbf{\hat{e}}) = cos(f_3(e), \frac{1}{K} \sum_{e' \in \mathbf{\hat{e}}}^{} f_3(e'))}
\end{equation}

\noindent \textcolor{blue}{where $K$ is the number of entities in the entity context $\mathbf{\hat{e}}$. In addition, three surface form similarity scores are leveraged by Yamada to capture  the similarity between the surface forms of the mention $m$ and the candidate entity $e$. They are the edit distance, whether the surface form of the candidate entity exactly equals or contains the mention, and whether the surface form of the candidate entity starts or ends with the mention, which are denoted by $\Psi^{(Y)}_{4}(e, m)$, $\Psi^{(Y)}_{5}(e, m)$ and $\Psi^{(Y)}_{6}(e, m)$, respectively. Based on these six scores, Yamada calculates the prediction score $g(e; m)$ for each candidate entity $e \in E(m)$ as follows:}

\begin{equation}
\begin{aligned}
    \textcolor{blue}{g(e; m) = \Gamma^{(Y)}(}
    & \textcolor{blue}{\Psi^{(Y)}_{1}(e, m), \Psi^{(Y)}_{2}(e, c), \Psi^{(Y)}_{3}(e, \mathbf{\hat{e}}),} \\
    & \textcolor{blue}{\Psi^{(Y)}_{4}(e, m), \Psi^{(Y)}_{5}(e, m),  \Psi^{(Y)}_{6}(e, m))}
\end{aligned}
\end{equation}

\noindent \textcolor{blue}{where $\Gamma^{(Y)}$ is a fully connected layer that generates the final score.}

% function that sequentially applies the following neural layers: [Linear(3, 100), Dropout(0.1), ReLU(), Linear(100, 1)] and results in the final score.

\subsubsection{{DeepED}}
\label{section:deeped}

{DeepED \cite{ganea2017deep} is a remarkable work which embeds entities and words in a common vector space and leverages a neural attention mechanism over local context windows.}

{We adopt an extended variation of the local EL model proposed in DeepED as the EL model, which utilizes four kinds of scores to perform linking. The first one is the prior probability $\Psi^{(D)}_{1}(e, m) = f_1(e, m)$. The second score, i.e., the context similarity score, is obtained via applying an attention mechanism to give different attention weights to the words in the context. Similar to Yamada, a word in the vocabulary $V$ is mapped to its embedding via a feature function $f^{(D)}_4(w):V \rightarrow \mathbb{R}^Q$. We define the context similarity score $\Psi^{(D)}_{2}(e, c)$ as follows:}

\begin{equation}
    {\Psi^{(D)}_{2}(e, c) = \sum_{w \in c} \alpha(w) \cdot f^\intercal_3(e) \phi_1 f^{(D)}_4(w)}
\end{equation}

\noindent {where $\phi_1$ is a diagonal matrix and $\alpha(w)$ is the attention weight of word $w$. Before defining $\alpha(w)$, we first introduce a support score $u(w)$ as follows:}

\begin{equation}
    {u(w) = \max_{e' \in E(m)} f^\intercal_3(e') \phi_2 f^{(D)}_4(w)}
\end{equation}

\noindent {where $\phi_2$ is another diagonal matrix. We identify the top $H' \leq H$ words with respect to the support score $u(w)$, and the set of these selected words is denoted by $c'$, and then the attention weight is defined as:}
\begin{equation}
    {\alpha(w) =
    \left\{
    \begin{array}{cl}
        \frac{\mbox{exp}(u(w))}{\sum_{w' \in c'} \mbox{exp}(u(w')) } & \mbox{for~} w \in c' \\
        0 & \mbox{otherwise}
    \end{array}
    \right.}
\end{equation}

\noindent {The third score is the topical coherence $\Psi^{(D)}_{3}(e, \mathbf{\hat{e}})$ based on entity embeddings, which is defined as follows:}

\begin{equation}
    {\Psi^{(D)}_{3}(e, \mathbf{\hat{e}})= \frac{1}{K} \sum_{e' \in \mathbf{\hat{e}}} (f_3^\intercal (e) \phi_3 f_3(e'))}
\end{equation}

\noindent {where $\phi_3$ is a diagonal matrix as well. The last score is defined as $\Psi^{(D)}_{4}(e, \mathbf{\hat{e}})= \frac{1}{K} \sum_{e' \in \mathbf{\hat{e}}}f_2(e, e')$, which captures the topical coherence between entities in a document based on relatedness scores introduced in Section \ref{sec:entitylinkingfeature}. DeepED also introduces a fully connected layer $\Gamma^{(D)}$ to generate the prediction score $g(e; m)$ for each candidate entity $e \in E(m)$, which is defined as follows:}

\begin{equation}
    {g(e; m) = \Gamma^{(D)}(\Psi^{(D)}_{1}(e, m), \Psi^{(D)}_{2}(e, c), \Psi^{(D)}_{3}(e, \mathbf{\hat{e}}), \Psi^{(D)}_{4}(e, \mathbf{\hat{e}}))}
\end{equation}
% \noindent \textcolor{blue}{where $\Gamma_2$ is a function that sequentially applies the following neural layers: [Linear(4, 100), Dropout(0.1), ReLU(), Linear(100, 1)], which is different from $\Gamma_1$.

%shen revision
\begin{figure*}[thb]
	\centering
	\includegraphics[width=1\textwidth]{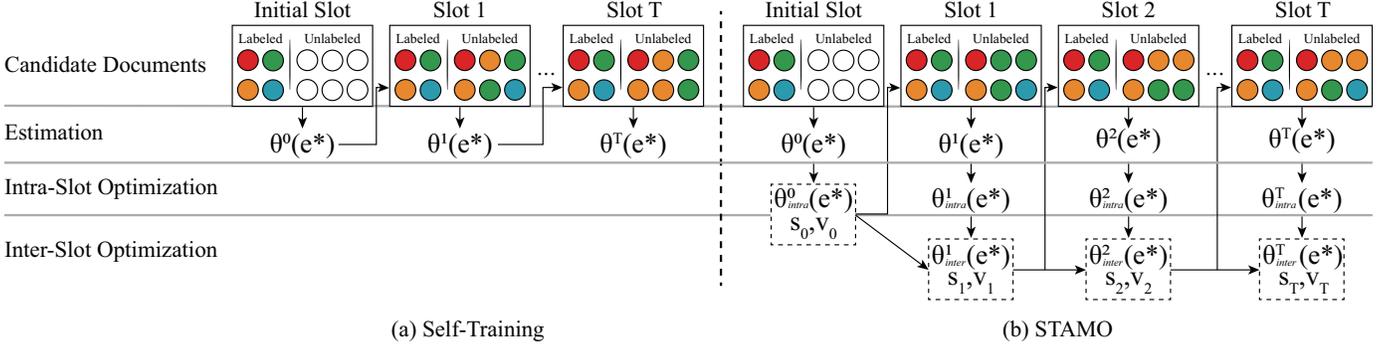}
%	\vspace{-2mm}
	\caption{The iterative process of \textbf{Self-Training} and our proposed \textbf{STAMO}. Circles of different colors represent different labels of candidate mentions in these candidate documents. 
		\textcolor{blue}{Note that $s_t$ and $v_t$ $(t= 0, 1, …, T)$ are the gradients (defined in Equation \ref{eq:sv}) used in inter-slot optimization.}}
	
	%Specifically, the label in the labeled candidate document is the \textbf{gold} corresponding entity of the candidate mention and remains unchanged in the iterative process. On the other hand, the label in the unlabeled candidate document is the \textbf{predicted} corresponding entity of the candidate mention, and may change in the iterative process. 
	
	\label{fig:approach}
	\vspace{-3mm}
\end{figure*}

\subsubsection{{Learning}}
\label{sec:learning}

\textcolor{blue}{The model parameters of Yamada only include the weights of the function $\Gamma^{(Y)}$,} while the model parameters of DeepED include the three diagonal matrices mentioned above (i.e., $\phi_1, \phi_2,$ and $\phi_3$) and the weights of the function $\Gamma^{(D)}$. The model parameters of these two introduced EL models are optimized with the same objective function $\mathcal{O}$ with a max-margin loss:

\begin{equation}
    {\mathcal{O}(\phi; S_{model}) = \sum_{(m, e^\dag) \in S_{model}} \sum_{e \in E(m)} [\mu - g(e^\dag; m) + g(e; m)]_+}
    \label{eq:of}
\end{equation}

% \begin{equation}
% \begin{split}
%     \textcolor{blue}{\mathcal{O}}&\textcolor{blue}{(\phi; S_{model}) =} \\
%        & \textcolor{blue}{\sum_{(m, e^\dag) \in S_{model}} \sum_{e \in E(m)} [\mu - s(e^\dag;m, c, \mathbf{\hat{e}}) + g(e;m, c, \mathbf{\hat{e}})]_+}
% \end{split}
% \end{equation}

\noindent {where $\phi$ denotes the model parameters of the EL model. $S_{model}$ is a labeled corpus for the EL model training and it only involves NEEs. In our experiments, $S_{model}$ is randomly collected from the annotated Web document corpus introduced in Section \ref{sec:wikipedia and web} and it contains 10K mention-entity pairs in total.}

\vspace{-1.4mm}
\section{The Proposed Approach: STAMO}

\subsection{Vanilla Self-Training}
\label{sec:vanilla self-training}

The simplest way to get feature parameters for EE $\theta(e^*)$ is just estimating them on the labeled data $L$, but we found when $|L|$ is small, simple estimation results in poor performance. Therefore, we try to leverage the knowledge hidden in the unlabeled data by self-training.

As shown in Figure \ref{fig:approach}(a), we initially estimate the EL features of EE $e^*$ on the labeled data $L$ and get $\theta^0(e^*)$ \textcolor{blue}{(w.r.t. line 1 in Algorithm \ref{algo})}. Next, we could link the candidate mentions to their mapping entities in the unlabeled data $U$ by using an EL model \textcolor{blue}{$g_{\theta^0(e^*)}$ (w.r.t. line 5 in Algorithm \ref{algo})}. Then, we estimate the EL features of EE $e^*$ on both real labeled and pseudo labeled data again and get $\theta^1(e^*)$ \textcolor{blue}{(w.r.t. line 6 in Algorithm \ref{algo})}. We repeat these steps until convergence and finally get $\theta^T(e^*)$ as the best result $\hat{\theta}(e^*)$ \textcolor{blue}{(w.r.t. line 15 in Algorithm \ref{algo})}. However, we found this vanilla self-training method did not work well either due to the error reinforcement problem. Specifically, in each slot of self-training, we link the candidate mentions in the unlabeled data $U$ to their mapping entities using an EL model. In most cases, the candidate mention is ambiguous and has more than one candidate entity. Some mapping entity of the candidate mention output by the EL model might be incorrect in reality, which makes some of the pseudo labels over the unlabeled data $U$ incorrect. In the next slot, the quality of EEs’ EL features estimated over these mislabeled data becomes worse and subsequently the linking result based on these learned EL features also becomes worse. Accordingly, this leads to the error reinforcement problem.

As shown in Figure \ref{fig:approach}(b), in \textbf{STAMO} we regard self-training as a multiple optimization process with respect to $\theta(e^*)$, and propose both intra-slot optimization and inter-slot optimization to alleviate the error reinforcement problem implicitly, which are introduced in the following.

\subsection{Intra-Slot Optimization}
\label{sec:intra-slot}

In each slot of self-training, we try to find the best feature parameters of EE $e^*$ that could minimize the objective function $\mathcal{O}$ (which was used to optimize the model parameters $\phi$ defined in Equation \ref{eq:of}) on the labeled data $L$ as follows:

\begin{equation}
	\theta_{intra}^t(e^*) = \argmin_{\theta(e^*)} \mathcal{O}(\theta(e^*); L)
	\label{eq:intra}
\end{equation}

\noindent and we consider $\theta^t(e^*)$, the estimated EL feature parameters of EE $e^*$ on both real labeled and pseudo labeled data, as an initial guess of this optimization process rather than the final result \textcolor{blue}{(w.r.t. lines 2 and 7 in Algorithm \ref{algo})}.
Intra-slot optimization could help the EL feature parameters of EE $e^*$ to be better combined with the optimal EL model parameters $\hat{\phi}$ and the optimal EL feature parameters of non-emerging entities $\hat{\theta}(\overline{e^*})$ estimated on Wikipedia, both of which are well trained before the EE is discovered and are frozen all the time when learning the EL feature parameters for the EE. This capacity is essential since we want to learn the EL features of EE $e^*$ incrementally.

\subsection{Inter-Slot Optimization}
\label{section:optimization}

Moreover, we try to optimize the EL features of EE $e^*$ by leveraging the information provided by the historical slots in self-training. To have this capacity, let us consider how $\theta_{intra}^t(e^*)$ is generated given the optimal EL feature parameters of the previous slot $\hat{\theta}^{t-1}(e^*)$. The \textit{real procedure} is: we use the EL model to link candidate mentions in $U$ with feature parameters $\hat{\theta}^{t-1}(e^*)$, and then we learn $\theta_{intra}^t(e^*)$ on both real labeled and pseudo labeled data as proposed in Section \ref{sec:intra-slot}. Instead of it, we construct a \textit{hypothetical procedure} which makes a direct connection between $\theta_{intra}^t(e^*)$ and $\hat{\theta}^{t-1}(e^*)$. We first rewrite $\theta_{intra}^t(e^*)$ by an identical equation:

\vspace{-2mm}
\begin{equation}
	\theta_{intra}^t(e^*) \equiv \hat{\theta}^{t-1}(e^*) - 1 \cdot (\hat{\theta}^{t-1}(e^*) - \theta_{intra}^t(e^*))
	\label{eq:rewrite}
\end{equation}

\noindent This identical equation can be explained as follows: consider a hypothetical objective function with respect to $\theta(e^*)$ whose analytical solution of derivative is unknown, but the gradient at the point $\hat{\theta}^{t-1}(e^*)$ is known, whose value is $\delta^t = \hat{\theta}^{t-1}(e^*) - \theta_{intra}^t(e^*)$ \textcolor{blue}{(w.r.t. line 8 in Algorithm \ref{algo})}. With this hypothetical objective function, we apply the gradient descent method at the point $\hat{\theta}^{t-1}(e^*)$ and set the learning rate to $\eta = 1$. According to the update rule of gradient descent, the updated point should be $\hat{\theta}^{t-1}(e^*) - \eta \cdot \delta^t$, which is equal to the right part of Equation \ref{eq:rewrite}. Therefore, we finally get $\theta_{intra}^t(e^*)$ after the update step. That is to say, in the hypothetical procedure, $\theta_{intra}^t(e^*)$ is generated by an inter-slot optimization process which has an update step for the feature parameters of $e^*$ in the previous slot, rather than the real intra-slot optimization process.

This novel perspective of self-training enables us to improve the EL features of EEs by leveraging the information of historical slots. Specifically, we adjust the gradient $\delta^t = \hat{\theta}^{t-1}(e^*) - \theta_{intra}^t(e^*)$ in consideration of the historical gradients, i.e., its moment estimates. First, the exponential moving averages of the gradient $s_t$ and the squared gradient $v_t$ are iteratively updated \textcolor{blue}{(w.r.t. line 9 in Algorithm \ref{algo})} as follows:
\begin{equation}
	\begin{split}
		s_t & = \beta_1 \cdot s_{t-1} + (1-\beta_1) \cdot \delta^t \\
		v_t & = \beta_2 \cdot v_{t-1} + (1-\beta_2) \cdot (\delta^t)^2
	\end{split}
	\label{eq:sv}
\end{equation}
\noindent where $\beta_1$ and $\beta_2$ are hyper-parameters that control the exponential decay rates and the initial values of $s_t$ and $v_t$ are both $\mathbf{0}$ \textcolor{blue}{(w.r.t. line 3 in Algorithm \ref{algo})}. Then the bias-corrected estimates are derived \textcolor{blue}{(w.r.t. line 10 in Algorithm \ref{algo})} as follows:
\begin{equation}
	\hat{s}_t = s_t/(1-\beta^t_1) \quad
	\hat{v}_t = v_t/(1-\beta^t_2)
	\label{eq:unbiasdsv}
\end{equation}
\noindent where $\beta^t$ denotes $\beta$ to the power $t$. Finally, the adaptive gradient is defined as $\hat{\delta}^t = \hat{s}_t/(\sqrt{\hat{v}_t}+\epsilon)$ \textcolor{blue}{(w.r.t. line 11 in Algorithm \ref{algo})}, where $\epsilon$ is a hyper-parameter to avoid division by zero error. This adjustment method for gradients is firstly proposed by Adam \cite{diederik2015adam} for the gradient descent method. Besides the adjustment of gradients, we further stabilize inter-slot optimization by decreasing the learning rate $\eta$ from $1$ to a small value, and we use a warm-up scheduling which increases the learning rate linearly in the first $\gamma$ slots. Therefore, the learning rate in the $t$-th slot is defined as $\eta^t = \textcolor{blue}{\min}(\eta, \eta \cdot t / \gamma)$ \textcolor{blue}{(w.r.t. line 12 in Algorithm \ref{algo})}. The final EL feature parameters of $e^*$ with the improved inter-slot optimization process are defined as follows:
\begin{equation}
\theta_{inter}^{t}(e^*) = \hat{\theta}^{t-1}(e^*) - \eta^t \cdot \hat{\delta}^t
\end{equation}

\noindent and it is also the optimal EL feature parameters of EE $e^*$ in the $t$-th slot (i.e., $\hat{\theta}^{t}(e^*) = \theta_{inter}^{t}(e^*)$) \textcolor{blue}{(w.r.t. line 13 in Algorithm \ref{algo})}. The pseudo code of our approach is depicted in Algorithm \ref{algo}.

%%%%%%%%%%%%%%%%%%%%%%%%%%%%%%%%%%%%%%%%%%%%%%%%%%%%%%%%%%%%%%%%%
%% Algorithm: STAMO
\begin{algorithm}[t]
\caption{\textbf{STAMO} (Self-Training as Multiple Optimizations)}
\begin{algorithmic}[1]
\INPUT Labeled set $L$, unlabeled set $U$, entity linking model $g$, learning rate $\eta$, warm-up period $\gamma$
\OUTPUT Optimal feature parameters $\hat{\theta}(e^*)$
% \Procedure{AdaptiveST}{$L, U, g, \eta, \gamma$}
% \State $\hat{\theta}^{0}(e^*) \leftarrow$ \textproc{FeatureLearning}($L, L, g$)
\State Estimate $\theta^{0}(e^*)$ on $L$
\State $\hat{\theta}^{0}(e^*) \leftarrow \theta_{intra}^{0}(e^*) \leftarrow$ intra-slot optimization with the initial guess $\theta^{0}(e^*)$ according to Equation \ref{eq:intra}
\State $s_0, v_0 \leftarrow 0$
\For{$t = 1,2,..., T$}
	\State Apply $g$ with $\hat{\theta}^{t-1}(e^*)$ to all unlabeled candidate mentions in $U$
	\State Estimate $\theta^{t}(e^*)$ on $L \cup \{(m, \hat{e}); m \in U\}$
	\State $\theta_{intra}^{t}(e^*) \leftarrow$ intra-slot optimization with the initial guess $\theta^{t}(e^*)$ according to Equation \ref{eq:intra}
	\State $\delta^t \leftarrow \hat{\theta}^{t-1}(e^*) - \theta_{intra}^{t}(e^*)$
	\State Compute $s_t, v_t$ according to Equation \ref{eq:sv}
	\State Compute $\hat{s}_t, \hat{v}_t$ according to Equation \ref{eq:unbiasdsv}
	\State $\hat{\delta}^t \leftarrow \hat{s}_t/(\sqrt{\hat{v}_t}+\epsilon)$ % \Comment{adaptive adjustment}
	\State $\eta^t \leftarrow \textcolor{blue}{\min}(\eta, \eta \cdot t / \gamma)$ % \Comment{warm up}
	\State $\hat{\theta}^{t}(e^*) \leftarrow \theta_{inter}^{t}(e^*) \leftarrow \hat{\theta}^{t-1}(e^*) - \eta^t \cdot \hat{\delta}^t$
\EndFor
\State \textbf{return} $\hat{\theta}^{T}(e^*)$
% \EndProcedure
\end{algorithmic}
\label{algo}

\end{algorithm}
%%%%%%%%%%%%%%%%%%%%%%%%%%%%%%%%%%%%%%%%%%%%%%%%%%%%%%%%%%%%%%%%%

%%%%%%%%%%%%%%%%%%%%%%%%%%%%%%%%%%%%%%%%

\textcolor{blue}{
We analyze the time and space complexity for Algorithm \ref{algo}.
The time complexity depends primarily on the iteration part (i.e., lines 4--14). At each iteration, the time complexity of line 5 is $O(K_l \cdot |U|)$, where $K_{l}$ is the cost to apply the EL model $g$ to one document and $|U|$ is the number of documents in $U$; the time complexity of line 6 is $O(K_e \cdot (|L|+|U|))$, where $K_{e}$ is the cost to estimate feature parameters with one document and $|L|$ is the number of documents in $L$; the time complexity of intra-slot optimization (i.e., line 7) is $O(K_{f} \cdot |L|)$, where $K_{f}$ is the cost to optimize feature parameters with one document; and the time complexity of inter-slot optimization (i.e., lines 8--13) is $O(1)$.
Therefore, the time complexity of the whole algorithm is $O(T \cdot (K_l \cdot |U| + K_e \cdot (|L|+|U|) + K_f \cdot |L|))$. As the size of the unlabeled data is far larger than that of the labeled data (i.e., $|U| >> |L|$), it could be simplified as $O(T \cdot (K_l+K_e)\cdot |U|)$.
The space complexity of the algorithm is $O(d)$, where $d$ is the sum of the dimensions of the learned EL features.
}

\subsection{STAMO Instantiations}

It is noted that \textbf{STAMO} is an approach of learning EL features for EEs based on self-training, which makes it flexibly integrated with any numerical EL feature or EL model. Depending on which specific EL features and EL models we choose to integrate with our approach \textbf{STAMO}, the instantiated whole EL systems may have different architectures and characteristics. However, these variations of instantiated whole EL systems are not the focus of this paper, as our interest is to provide an approach of EL feature learning for EEs rather than enumerating all the possibilities of combinations for the whole EL system. In the following experiment, we choose the two EL models (i.e., Yamada introduced in Section \ref{section:yamada} and DeepED introduced in Section \ref{section:deeped}) and the EL features they utilize to work with our \textbf{STAMO}.

\section{Experiments}
% To evaluate the effectiveness of our proposed approach \textbf{STAMO}, we conduct a thorough experimental study, and present the experimental results and analyses in this section. We firstly describe the experimental setting in Section \ref{sec:experimental settings} and then study the effectiveness of our approach \textbf{STAMO} in comparison with several baselines in Section \ref{sec:effectiveness study}. In Section \ref{sec:convergence study}, we present a convergence study of self-training. In Section \ref{sec:ablation study}, we conduct two  ablation studies. Finally, we investigate how the labeled data size influences the quality of the learned EL features for EEs in Section \ref{sec:effect analysis}.
% In this section, we empirically verify the effectiveness of our proposed approach.  We compare our approach with various baseline methods on two datasets. Then we perform a convergence study and an ablation study to further investigate the effects of different modules applied in our approach. Finally, we target at answering the following question: how does the number of labeled documents influence the final quality of EEs' features? The datasets and codes are public available\footnote{\url{https://anonymous.4open.science/r/f0485db8-edd0-4510-958f-034f58ecbcdc/}}.

\subsection{Experimental Setting}
\label{sec:experimental settings}
To the best of our knowledge, there is no publicly available benchmark dataset for the task of learning entity linking features for emerging entities. According to the task definition introduced in Section \ref{sec:Preliminaries}, for each EE, a partially labeled Web document corpus $W$ should be given as the input for learning its EL features. $W$ is a reasonable number of candidate documents each of which contains at least one candidate mention w.r.t. the EE, and a small subset of $W$ should be manually labeled, i.e., the gold corresponding entities for candidate mentions in documents are annotated. However, all the publicly available entity linking datasets (e.g., AIDA-CoNLL \cite{hoffart2011robust}, TAC-KBP2010 \cite{ji2011knowledge}, ACE2004 \cite{ratinov2011local}, MSNBC \cite{cucerzan2007large}, AQUAINT \cite{milne2008learning} and so on) are constructed for the case of linking with non-emerging entities, and do not have a partially labeled Web document corpus for each entity as each entity is mentioned just a few times (mostly once or twice) in these datasets. This limited occurrence of the entity in these datasets cannot provide adequate context for EL feature learning. Accordingly, these publicly available EL datasets do not meet the requirements of our new task, and are unsuitable for the evaluation of our new task of learning EL features for EEs. Therefore, we create a partially labeled Web document corpus $W$ for each EE and additionally construct two EL test datasets involving the selected EEs for our new task. We make the datasets and the source code used in this paper publicly available for future research\footnote{\url{https://github.com/stamo4el/STAMO}}.

\subsubsection{Emerging Entities}
\label{sec:emerging entities}
Since collecting and annotating data for \emph{real} EEs are difficult, we select some entities following the criteria proposed in the prior work \cite{singh2016discovering} and regard them as EEs. The first criterion is that the selected entity should not occur very frequently. On the other hand, a reasonable number of candidate documents for this entity is also required, as we need to create a partially labeled Web document corpus $W$ for this entity. Therefore, we bound the document frequency of the selected entity in the range of 1000 to 2000. The second criterion is \emph{ambiguous}. An alias is considered as ambiguous if it could refer to more than one entity. We provide the top-3 most popular aliases for each selected entity, and we require them all to be ambiguous in order to make the EL problem challenging. Otherwise, if an alias of an entity is unambiguous, it is effortless to link this alias mentioned in text with the only mapping entity correctly. We further define the ambiguity rate of an entity as the proportion of candidate mentions that really refer to it in all its candidate mentions. We bound the ambiguity rate of the selected entity in the range of 0.2 to 0.8, as we hope a reasonable proportion of candidate mentions refer to the selected entity, not too much or too little. Finally, we randomly select 50 entities meeting the above criteria and regard them as EEs. Their average ambiguity rate is $0.61$. We initialize the feature parameters associated with them to zero, and the only information known about them is at most 3 aliases.

\subsubsection{Wikipedia and Web Documents}
\label{sec:wikipedia and web}

The august 2017 version of English Wikipedia dump is used in our experiments. It contains about 5.5 million article pages, each of which contains 14.2 hyperlinks (i.e., gold mention-entity pairs) on average. The hyperlinks in Wikipedia are expressed by handcraft internal links using wikitext markup\footnote{\url{https://en.wikipedia.org/wiki/Help:Wikitext}} and could be identified easily. Wikipedia is used to estimate EL features for non-emerging entities (NEEs), and it is also the target KB in our experiment.

Following the prior work \cite{singh2016discovering}, the ClueWeb12-B13 dataset\footnote{\url{http://lemurproject.org/clueweb12}} is chosen as the source of Web documents in our experiments, and it contains more than 52 million Web documents in total. The FACC1 dataset \cite{gabrilovich2013facc1} provides annotations of mention-entity pairs for the Web documents in ClueWeb by an automatic EL method, and the annotation precision is believed to be around 80-85\%. Despite that the annotation provided by the FACC1 dataset is not perfect, it is the largest and the most suitable dataset providing the EL annotation for the Web documents in ClueWeb to the best of our knowledge. Each Web document in ClueWeb has 13 mentions annotated on average.
% It is believed that the precision and recall of these automatic annotations are around 80-85\% and 70-85\% respectively.

For each EE, we randomly sampled 1000 candidate documents each of which contains at least one candidate mention of the EE from ClueWeb, and regarded them as the Web document corpus $W$. Then we leverage $W$ as the source to learn EL features for the EE (i.e., $|W|=1000$), among which only 20 candidate documents were labeled (i.e., $|L|=20$). The labels of the candidate mentions in this labeled subset $L$ are provided by FACC1, while the labels of the candidate mentions in the remaining unlabeled set $U$ remain unknown.

\subsubsection{EL Datasets}
After learning EL features for the EE based on the Web document corpus $W$, the next step is to evaluate the quality of the learned EL features. Directly evaluating the feature quality is difficult, as we cannot obtain the gold standard EL feature parameters for emerging entities. Thus, we evaluate the EL feature quality via the target downstream task (i.e., entity linking). Specifically, we evaluate the quality of the learned EL features, via assessing the entity linking result of some EL model that leverages the learned EL features for linking candidate mentions of the selected EEs. Therefore, we need to construct EL datasets containing candidate mentions of the selected EEs.

We first construct the EL dataset \textbf{Test-Web} from ClueWeb. It is comprised of 200 reserved candidate documents in ClueWeb (i.e., the documents not appearing in $W$) for each of the 50 selected EEs, and thus this EL dataset \textbf{Test-Web} contains 10K documents in total.
% We construct two EL datasets to evaluate different approaches: \textbf{Test-Web} and \textbf{Test-Wiki}. They are comprised of 200 reserved candidate (i.e., the documents not appearing in $W$) documents in ClueWeb12-B13 and Wikipedia respectively for each of the 50 selected EEs, and thus each of them contains 10K documents in total.

%\begin{itemize}
%    \item{\textbf{Test-Web}:} it is comprised of 200 reserved candidate documents in ClueWeb12-B13 (i.e., the documents not appearing in $W$) for each of the 50 selected EEs, and thus it contains 10K documents in total.
%    \item{\textbf{Test-Wiki}:} it is comprised of 200 candidate documents in Wikipedia for each EE and thus it also contains 10K documents in total.
% \end{itemize}

\begin{table}[t]
%\begin{mdframed}[linecolor=blue,linewidth=1pt,innerrightmargin=6pt,innerbottommargin=6pt,innerleftmargin=6pt,innertopmargin=2pt,backgroundcolor=white]
		\centering
		\caption{Statistics of the constructed datasets.}
			\vspace{-2mm}
		\label{tab:datasets}
		\scalebox{1}{
			\begin{tabular}{lccc}
				\toprule
				\multirow{2}{*}{Dataset}                                                 & Web document & EL dataset        & EL dataset         \\
				&  corpus \boldmath{$W$} \unboldmath   & \textbf{Test-Web} & \textbf{Test-Wiki} \\ \midrule
				\begin{tabular}[c]{@{}c@{}}\# candidate \\ documents\end{tabular}        & 50,000       & 10,000            & 10,000             \\ \midrule
				\begin{tabular}[c]{@{}c@{}}\# candidate \\ documents \\per EE\end{tabular} & 1,000        & 200               & 200                \\
				\bottomrule
		\end{tabular}}
%	\end{mdframed}
	\vspace{-4mm}
%\vspace{-3.5mm}
\end{table}

Since the labels of the candidate mentions in \textbf{Test-Web} are generated by an automatic EL method and thus imperfect, we construct the second EL dataset {\textbf{Test-Wiki}. It is comprised of 200 candidate documents in Wikipedia for each of the 50 selected EEs and thus it also contains 10K documents in total. In this \textbf{Test-Wiki} dataset, the labels of the candidate mentions are generated using hyperlinks in Wikipedia, which are usually correct annotations. Therefore, we consider that the evaluation results on \textbf{Test-Wiki} are more convincing compared with \textbf{Test-Web}. Additionally, the \textbf{Test-Wiki} dataset is helpful for us to verify if the learned EEs' features can generalize well to different data sources, as EL features are learned based on the Web documents from ClueWeb. \textcolor{blue}{The statistics of the constructed datasets are shown in Table \ref{tab:datasets}.}

\subsubsection{Baselines and Metrics}

\begin{table*}[t]
	\caption{Experimental results over the two EL datasets}
		\vspace{-2mm}
	\begin{tabularx}{1\textwidth}{X *{12}{s}}
		
		\toprule
		
		\multirow{3}{*}{Methods} & & \multicolumn{4}{c}{\textbf{Test-Web}} & \multicolumn{4}{c}{\textbf{Test-Wiki}} & \multicolumn{2}{|c|}{Avg.} & \multirow{3}{*}{\makecell{Time\\(min)}} \\
		\cmidrule(lr){3-6} \cmidrule(lr){7-10} \cmidrule(lr){11-12} & &
		Acc  &  P  &  R   &  $\mbox{F}_1$  & Acc   &  P  &  R   &  \multicolumn{1}{c|}{$\mbox{F}_1$}  &  Acc  &  \multicolumn{1}{c|}{$\mbox{F}_1$}  & \\
		
		\midrule
		
		\textbf{Keyphrase}                                                                     &                                                                           & 33.32                            & 82.56                     & 48.83                     & 61.39                            & 44.96                            & 87.15                     & 65.23                     & \multicolumn{1}{c|}{74.61}                            & 39.14                            & \multicolumn{1}{c|}{68.00}                            & - \\
		\textcolor{blue}{\textbf{BERT}}                                                        &                                                                           & \textcolor{blue}{53.88}          & \textcolor{blue}{{98.24}} & \textcolor{blue}{29.51}   & \textcolor{blue}{45.39}          & \textcolor{blue}{52.44}          & \textcolor{blue}{{95.53}} & \textcolor{blue}{38.32}   & \multicolumn{1}{c|}{\textcolor{blue}{54.70}}          & \textcolor{blue}{53.16}          & \multicolumn{1}{c|}{\textcolor{blue}{50.04}}          & - \\
		
		\midrule
		
		\multicolumn{1}{l|}{\textcolor{blue}{\textbf{Estimation}}}                             & \multicolumn{1}{c|}{\multirow{6}{*}{\textcolor{blue}{\textbf{+ Yamada}}}} & \textcolor{blue}{46.03}          & \textcolor{blue}{88.75}   & \textcolor{blue}{45.03}   & \textcolor{blue}{59.75}          & \textcolor{blue}{43.03}          & \textcolor{blue}{92.43}   & \textcolor{blue}{26.24}   & \multicolumn{1}{c|}{\textcolor{blue}{40.87}}          & \textcolor{blue}{44.53}          & \multicolumn{1}{c|}{\textcolor{blue}{50.31}}          & - \\
		\multicolumn{1}{l|}{\textcolor{blue}{\textbf{Estimation(1K)}}}                         & \multicolumn{1}{c|}{}                                                     & \textcolor{blue}{\textbf{65.39}} & \textcolor{blue}{87.77}   & \textcolor{blue}{78.76}   & \textcolor{blue}{\textbf{83.02}} & \textcolor{blue}{59.16}          & \textcolor{blue}{88.59}   & \textcolor{blue}{57.74}   & \multicolumn{1}{c|}{\textcolor{blue}{69.92}}          & \textcolor{blue}{62.27}          & \multicolumn{1}{c|}{\textcolor{blue}{76.47}}         & - \\
		\multicolumn{1}{l|}{\textcolor{blue}{\textbf{Self-Training}}}                          & \multicolumn{1}{c|}{}                                                     & \textcolor{blue}{47.07}          & \textcolor{blue}{71.97}   & \textcolor{blue}{57.17}   & \textcolor{blue}{63.72}          & \textcolor{blue}{48.93}          & \textcolor{blue}{80.70}   & \textcolor{blue}{41.62}   & \multicolumn{1}{c|}{\textcolor{blue}{54.92}}          & \textcolor{blue}{48.00}          & \multicolumn{1}{c|}{\textcolor{blue}{59.32}}          & 13.52 \\
		\multicolumn{1}{l|}{\hspace{0.2cm}+\textcolor{blue}{\textit{Intra-Slot Optimization}}} & \multicolumn{1}{c|}{}                                                     & \textcolor{blue}{62.00}          & \textcolor{blue}{79.92}   & \textcolor{blue}{{79.44}} & \textcolor{blue}{79.68}          & \textcolor{blue}{68.13}          & \textcolor{blue}{91.20}   & \textcolor{blue}{{72.43}} & \multicolumn{1}{c|}{\textcolor{blue}{80.74}}          & \textcolor{blue}{65.06}          & \multicolumn{1}{c|}{\textcolor{blue}{80.21}}         & 13.82 \\
		\multicolumn{1}{l|}{\hspace{0.2cm}+\textcolor{blue}{\textit{Inter-Slot Optimization}}} & \multicolumn{1}{c|}{}                                                     & \textcolor{blue}{51.42}          & \textcolor{blue}{87.44}   & \textcolor{blue}{55.47}   & \textcolor{blue}{67.88}          & \textcolor{blue}{49.48}          & \textcolor{blue}{91.41}   & \textcolor{blue}{37.77}   & \multicolumn{1}{c|}{\textcolor{blue}{53.46}}          & \textcolor{blue}{50.45}          & \multicolumn{1}{c|}{\textcolor{blue}{60.67}}         & 13.63 \\
		\multicolumn{1}{l|}{\textcolor{blue}{\textbf{STAMO}}}                                  & \multicolumn{1}{c|}{}                                                     & \textcolor{blue}{63.86}          & \textcolor{blue}{89.95}   & \textcolor{blue}{76.28}   & \textcolor{blue}{82.55}          & \textcolor{blue}{\textbf{68.94}} & \textcolor{blue}{94.53}   & \textcolor{blue}{71.18}   & \multicolumn{1}{c|}{\textcolor{blue}{\textbf{81.21}}} & \textcolor{blue}{\textbf{66.40}} & \multicolumn{1}{c|}{\textcolor{blue}{\textbf{81.88}}} & 13.95 \\
		
		\midrule
		
		\multicolumn{1}{l|}{\textbf{Estimation}}                                               & \multicolumn{1}{c|}{\multirow{6}{*}{\textbf{+ DeepED}}}                   & 49.50                            & 78.87                     & 55.75                     & 65.32                            & 49.06                            & 94.23                     & 18.24                     & \multicolumn{1}{c|}{30.56}                            & 49.28                            & \multicolumn{1}{c|}{47.94}                            & - \\
		\multicolumn{1}{l|}{\textbf{Estimation(1K)}}                                           & \multicolumn{1}{c|}{}                                                     & 68.11                            & 81.41                     & 87.72                     & \textbf{84.45}                   & 64.43                            & 90.81                     & 47.79                     & \multicolumn{1}{c|}{62.63}                            & 66.27                            & \multicolumn{1}{c|}{73.54}                            & - \\
		\multicolumn{1}{l|}{\textbf{Self-Training}}                                            & \multicolumn{1}{c|}{}                                                     & 55.25                            & 63.50                     & 77.57                     & 69.83                            & 59.47                            & 86.40                     & 40.76                     & \multicolumn{1}{c|}{55.39}                            & 57.36                            & \multicolumn{1}{c|}{62.61}                           & 12.13 \\
		\multicolumn{1}{l|}{\hspace{0.2cm}+\textit{Intra-Slot Optimization}}                   & \multicolumn{1}{c|}{}                                                     & 63.32                            & 69.74                     & 87.27                     & 77.52                            & 81.85                            & 93.67                     & 77.40                     & \multicolumn{1}{c|}{84.76}                            & 72.54                            & \multicolumn{1}{c|}{81.14}                            & 12.60 \\
		\multicolumn{1}{l|}{\hspace{0.2cm}+\textit{Inter-Slot Optimization}}                   & \multicolumn{1}{c|}{}                                                     & 62.36                            & 66.42                     & 89.63                     & 76.30                            & 66.13                            & 88.33                     & 52.38                     & \multicolumn{1}{c|}{65.76}                            & 64.25                            & \multicolumn{1}{c|}{71.03}                            & 12.72 \\
		\multicolumn{1}{l|}{\textbf{STAMO}}                                                    & \multicolumn{1}{c|}{}                                                     & \textbf{68.48}                   & 70.06                     & {97.87}                   & \textbf{}81.66                   & \textbf{86.19}                   & 92.71                     & {86.03}                   & \multicolumn{1}{c|}{\textbf{89.24}}                   & \textbf{77.34}                   & \multicolumn{1}{c|}{\textbf{85.45}}                   & 13.25 \\
		
		\bottomrule
	\end{tabularx}
	\label{tab:main1}
	\vspace{-2mm}
\end{table*}
%%%%%%%%%%%%%%%%%%%%%%%%%%%%%%%%%%%%%%%%%%%%%%

We study the task of learning EL features for EEs, which is orthogonal to the task of developing an EL model. In order to evaluate different approaches of learning EL features for EEs, we assess the quality of their learned EL features, whereas the performance of different EL models is not the focus of this paper. Therefore, to demonstrate the effectiveness of our proposed approach \textbf{STAMO}, we compare it with different approaches of learning EL features for EEs, rather than different EL models.

To the best of our knowledge, there is only one prior work \cite{singh2016discovering} (called \textbf{Keyphrase}) addressing the task of learning EL features for EEs. In addition to it, we also create several baselines which are introduced in the following:

\begin{itemize}
	\item{\textbf{Keyphrase}:} the keyphrase-based method proposed in \cite{singh2016discovering}. Following the setting in their original paper, the keyphrase descriptions mined from Wikipedia for EEs are regarded as gold standard. We assume this method could successfully mine all the gold standard keyphrase descriptions for EEs and thus the result we report for this method could be regarded as the upper bound of this method. According to their setting, after mining keyphrase descriptions for EEs, candidate mentions in text could be linked based on context overlap with keyphrases via the EL model proposed in \cite{hoffart2011robust}, while our EL models (introduced in Section \ref{section:model}) cannot leverage the mined keyphrase descriptions of EEs well for entity linking.

    \item{\textcolor{blue}{\textbf{BERT}:}} \textcolor{blue}{the method that leverages BERT \cite{devlin2019bert} to learn representations for entities to perform entity linking. Following the same setting in existing BERT-based EL works \cite{logeswaran2019zero,wu2020scalable,tang2021bidirectional}, we employ a BERT-based cross-encoder to jointly encode the mention context and the entity description to perform cross-attention between each other. For each NEE, we could obtain its entity description from its corresponding Wikipedia page. Since EEs have not been included in Wikipedia yet, we form the entity description for each EE via concatenating contexts of mentions that are labeled as that EE from the labeled data $L$. For linking, the concatenation of the mention context and the entity description is taken as the input sequence of BERT and a linear layer is applied to the output representation of the \texttt{[CLS]} token to generate the ranking score for each candidate entity.
    	The parameters of BERT are initialized by the BERT base-uncased model in the experiment.
    	We fine-tune this method using the labeled data  $S_{model}$ for the EL model training introduced in Section \ref{sec:learning}, based on a cross-entropy loss, the same as existing BERT-based EL works \cite{logeswaran2019zero,wu2020scalable,tang2021bidirectional}.}

	\item{\textbf{Estimation}:} the method that simply estimates EEs' EL features based on the given labeled data $L$. To demonstrate the performance of simply estimating EL features over a large number of labeled candidate documents for each EE, we also tried to increase the size of $L$ from 20 to 1000 (i.e., 1K) and this method is denoted by \textbf{Estimation(1K)}.
    %This method could be regarded as the initial slot shown in Figure \ref{fig:approach}(a). Thus we could also extend it by adding intra-slot optimization to it, as the initial slot shown in Figure \ref{fig:approach}(b).
	\item{\textbf{Self-Training}:} the vanilla self-training method as introduced in Section \ref{sec:vanilla self-training}. It is noted that only 20 labeled candidate documents are provided for each EE in the \textbf{Self-Training} method. We also tried to improve it via adding intra-slot optimization or inter-slot optimization to it separately. These self-training based baselines are different variants of our proposed \textbf{STAMO}. Performance comparison of these baselines and \textbf{STAMO} could be regarded as an ablation study to investigate the contribution of each component in \textbf{STAMO}.
\end{itemize}

We assess the quality of the learned EL features for EEs via the entity linking result performed on candidate mentions of these EEs. {It is noted that all the baselines introduced above except \textbf{Keyphrase} and \textbf{BERT} are integrated with the same EL model} as our proposed approach \textbf{STAMO} to perform entity linking. Therefore, the different EL results of those baselines and our proposed \textbf{STAMO} are only determined by the EL features they learn. That is to say, the better the final EL result is, the better the quality of the learned EL features is.

The first metric we choose to assess the EL result is accuracy. Given some test candidate mentions whose gold entities are $\{e^\dag_1,...,e^\dag_N\}$ and predicted entities are $\{\hat{e}_1,...\hat{e}_N\}$, accuracy is defined as follows:
\begin{equation*}
	Acc= \frac{\sum_{i=1}^N \mathbbm{1}(\hat{e}_i = e^{\dag}_i)}{N}
\end{equation*}
\noindent where $\mathbbm{1}(x)$ is an indicator function whose value is $1$ if condition $x$ is true, otherwise the value is $0$. Since in this paper we focus on EEs, we further consider an EE as a query, and define precision, recall, and $\mbox{F}_1$-score with respect to the EE as follows:
\begin{gather*}
	P = \frac{\sum_{i=1}^N \mathbbm{1}(\hat{e}_i = e^{\dag}_i \mbox{~and~} \hat{e}_i = e^*)}{\sum_{i=1}^N \mathbbm{1}(\hat{e}_i = e^*)} \\
	R = \frac{\sum_{i=1}^N \mathbbm{1}(\hat{e}_i = e^{\dag}_i \mbox{~and~} e^{\dag}_i = e^*)}{\sum_{i=1}^N \mathbbm{1}(e^{\dag}_i = e^*)} \quad
	\mbox{F}_1 = 2 \cdot \frac{P \cdot R}{P + R}
\end{gather*}

\subsubsection{Preprocessing}
For the ClueWeb12-B13 dataset, we extract main textual contents of Web documents by Boilerpipe\footnote{\url{https://boilerpipe-web.appspot.com/}}. For Wikipedia, we choose WikiExtractor\footnote{\url{https://github.com/attardi/wikiextractor}} to preprocess the dump so that we can identify these mentions and their corresponding entities in Wikipedia articles. As the label of the mention in the Web document provided by the FACC1 dataset is its corresponding Freebase entity, we align the entity in Freebase to Wikipedia via the attribute \enquote{key:wikipedia.en\_id}.

\subsubsection{Training Details}
In our experiments, most hyper-parameters are set to the recommended value, and we only describe the different parts here. When updating entity embeddings of EEs, we train them for 35 epochs with the learning rate of 1e-2. For the EL model described in Section \ref{section:model}, we keep the top 30 candidate entities for each mention based on prior probability when training, while all candidate entities are kept when testing. \textcolor{blue}{The fully connected layer $\Gamma^{(Y)}$ used in Yamada sequentially applies the neural layers [Linear(6, 100), Dropout(0.1), ReLU(), Linear(100, 1)],} while the fully connected layer $\Gamma^{(D)}$ used in DeepED sequentially applies the neural layers [Linear(4, 100), Dropout(0.1), ReLU(), Linear(100, 1)]. In intra-slot optimization, we train $\theta(e^*)$ for 50 epochs with the learning rate of 1e-4, and the batch size is set to 32.
In inter-slot optimization, we apply different learning rates $\eta$ to different groups of feature parameters as follows: $\eta(\theta_1)=\mbox{5e-2}, \eta(\theta_2)=\mbox{5e-2}, \eta(\theta_3)=\mbox{1e-3}$. The warm-up period $\gamma$ is set to $5$.

\subsection{Effectiveness Study}
\label{sec:effectiveness study}
\vspace{-0.5mm}

The experimental results of all approaches over the two EL datasets (i.e., \textbf{Test-Web} and \textbf{Test-Wiki}) are shown in Table \ref{tab:main1}. From the results, it can be seen that even though the gold standard keyphrase descriptions are provided for EEs, \textbf{Keyphrase} does not perform well since keyphrase description is not a strong and effective EL feature. \textcolor{blue}{The baseline \textbf{BERT} also performs poorly on both datasets, which may be due to the reason that it is unable to learn high-quality representations for EEs as the absence of their corresponding Wikipedia pages. In addition to the two baselines above (i.e., \textbf{Keyphrase} and \textbf{BERT}), other approaches shown in Table 3 (e.g., \textbf{Estimation}, \textbf{Estimation(1K)}, \textbf{Self-Training}, and \textbf{STAMO}) are integrated with each of the two EL models (i.e., Yamada and DeepED) to perform entity linking. From the results, we can see that the performance of these approaches has a similar trend when they are integrated with Yamada and DeepED. It is also noted that we should compare their performance when they are integrated with the same EL model, either Yamada or DeepED. Specifically,} when 20 labeled candidate documents are provided for each EE, \textbf{Estimation} \textcolor{blue}{(+Yamada or +DeepED)} which simply estimates EEs' EL features on these labeled data does not perform well either. When the number of labeled candidate documents increases greatly to 1K, we could see that \textbf{Estimation(1K)} \textcolor{blue}{(+Yamada or +DeepED)} achieves the highest F1-score on \textbf{Test-Web}, which is consistent with our intuition. However, this method is impractical in reality because manually annotating such a large number (i.e., 1K) of candidate documents for each EE is very time-consuming and labor-intensive. To our surprise, \textbf{Estimation(1K)} \textcolor{blue}{(+Yamada or +DeepED)} does not achieve good performance on the \textbf{Test-Wiki} dataset. The reason may be that the underlying distributions of documents from  Wikipedia and Web are different. EL features simply estimated from Web documents in ClueWeb do not generalize well to Wikipedia documents in \textbf{Test-Wiki}.

\begin{figure*}[t]
	\centering
	\includegraphics[width=0.92\textwidth]{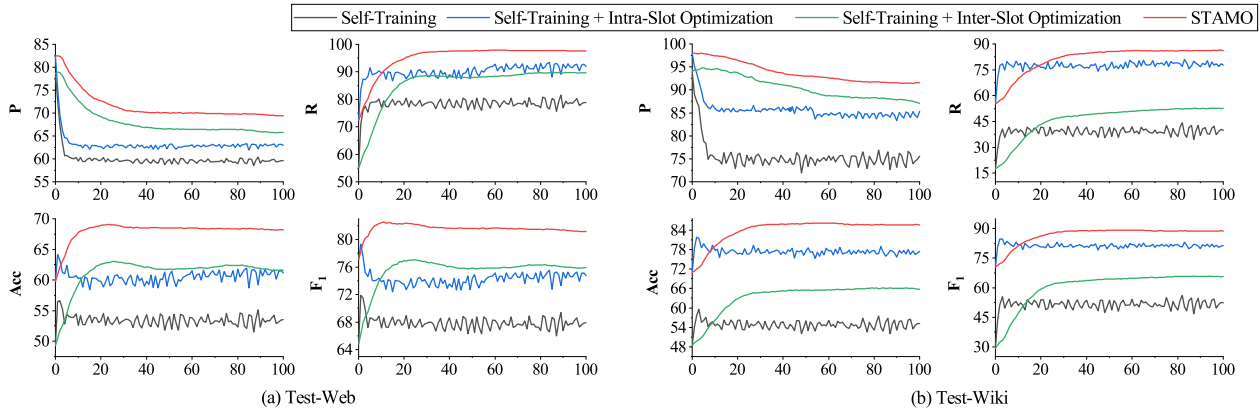}
		\vspace{-3mm}
	\caption{Performances of different self-training based methods as iterations progress. Horizontal axes represent the number of iterations $t$.
	}
	\label{fig:convergence}
	\vspace{-3mm}
\end{figure*}

\textcolor{blue}{Compared with \textbf{Estimation}, \textbf{Self-Training} with the same size of the labeled data $L$ achieves much better results. Specifically, when integrated with Yamada, \textbf{Self-Training} achieves improvements of 3.47\% average accuracy and 9.01\% average $\mbox{F}_1$-score compared with \textbf{Estimation}. When integrated with DeepED, \textbf{Self-Training} achieves improvements of 8.08\% average accuracy and 14.67\% average $\mbox{F}_1$-score compared with \textbf{Estimation}.  It indicates that \textbf{Self-Training} leverages the knowledge hidden in the unlabeled data effectively to learn better EL features.} Then, adding intra-slot optimization or inter-slot optimization to \textbf{Self-Training} \textcolor{blue}{(+Yamada or +DeepED)} separately can both improve this vanilla self-training method in terms of average accuracy and average $\mbox{F}_1$-score. Finally, with the use of both intra-slot optimization and inter-slot optimization, \textbf{STAMO} \textcolor{blue}{(+Yamada or +DeepED)} achieves the best performance in terms of average accuracy and average F1-score. \textcolor{blue}{Specifically, when integrated with Yamada, \textbf{STAMO} achieves improvements of 27.26\% (13.24\%) average accuracy and 13.88\% (31.84\%) average $\mbox{F}_1$-score compared with the baseline \textbf{Keyphrase} (\textbf{BERT}). When integrated with DeepED, \textbf{STAMO} achieves improvements of 38.20\% (24.18\%) average accuracy and 17.45\% (35.41\%) average $\mbox{F}_1$-score compared with the baseline \textbf{Keyphrase} (\textbf{BERT}).} Moreover, \textbf{STAMO} even outperforms \textbf{Estimation(1K)} on the \textbf{Test-Wiki} dataset and on average by a large margin, under the condition that \textbf{Estimation(1K)} uses a labeled data whose size is 50 times larger than \textbf{STAMO}. This demonstrates that our proposed \textbf{STAMO} could learn high-quality EL features for EEs with the requirement of just a small number of labeled documents for each EE and each component of \textbf{STAMO} has a significant positive contribution to the \textbf{STAMO}’s performance.

\textcolor{blue}{From the results in Table \ref{tab:main1}, we can also see that DeepED performs better than Yamada in most cases when they work with the same EL feature learning approach, as DeepED leverages a neural attention mechanism over local context windows and the topical coherence based on relatedness scores. However, as stated before, the performance of different EL models is not the focus of this paper since this paper focuses on the task of learning EL features for EEs rather than developing an EL model.}

The run-times of \textbf{STAMO} (and its several variants) and the baselines (i.e., \textbf{Keyphrase} and \textbf{BERT}) are not comparable, as they learn EL features in totally different ways. Therefore, we just demonstrate the run-times of our \textbf{STAMO} and its variants (i.e., the other three self-training based methods). As the goal of our task is to learn EL features for EEs based on the Web document corpus $W$, we show the average EL feature learning time per iteration for each self-training based method in Table \ref{tab:main1}. From the results, we can see that the vanilla self-training method \textbf{Self-Training} has the least time consumption among all the self-training based methods when integrated with Yamada or DeepED. Adding intra-slot or inter-slot optimization to \textbf{Self-Training} would increase a small amount of learning time. Finally, the complete method \textbf{STAMO} is the most time-consuming, which is consistent with our intuition. From another perspective, we can see that among the whole process of \textbf{STAMO}, the vanilla self-training process (including applying the EL model to the unlabeled data and estimating EL features based on both real labeled and pseudo labeled data) consumes most of the learning time, while the two optimization processes consume much less learning time, which is consistent with our time complexity analysis depicted in Section \ref{section:optimization}.

\subsection{Convergence Study}
\label{sec:convergence study}

Self-training does not guarantee to converge since early errors may be reinforced. In this experiment, we evaluate the performances of different self-training based methods as iterations progress. Their experimental results {based on DeepED} are shown in Figure \ref{fig:convergence}. We could see intra-slot optimization and inter-slot optimization are both helpful, while inter-slot optimization plays a critical role in the stabilization of the iterative process. Then, with the help of intra-slot optimization and inter-slot optimization together, \textbf{STAMO} achieves the best performance on the two EL datasets in terms of all the metrics among these self-training based methods.
% That is to say, simple self-training is very sensitive to the choice of maximum iteration number. On the other hand, adaptive self-training is much more stable and converges to better results on both test sets, which demonstrates our adaptive adjustment to self-training

\begin{table}[t]
	\centering
	\caption{Ablation study of inter-slot optimization}
	\vspace{-3mm}
	\begin{tabular}{l|cccc|cc}
	\toprule
	\multicolumn{1}{l}{\multirow{2}{*}{Methods}} & \multicolumn{2}{c}{\textbf{Test-Web}}         & \multicolumn{2}{c|}{\textbf{Test-Wiki}}       & \multicolumn{2}{c}{Avg.}         \\
	\cmidrule(l){2-3}\cmidrule(l){4-5}\cmidrule(l){6-7}
	\multicolumn{1}{l}{}                         & Acc            & $\mbox{F}_1$ & Acc            & $\mbox{F}_1$ & Acc            & $\mbox{F}_1$    \\
	\midrule
	\textbf{STAMO}                               & \textbf{68.48} & \textbf{81.66}               & \textbf{86.19} & \textbf{89.24}               & \textbf{77.34} & \textbf{85.45}  \\
	w/o changing $\delta$                        & 66.83          & 81.03                        & 84.09          & 87.05                        & 75.50          & 84.03           \\
	w/o changing $\eta$                          & 64.25          & 79.46                        & 85.31          & 89.05                        & 74.78          & 84.26           \\
	w/o warm-up                                  & 68.32          & 81.53                        & 85.66          & 88.89                        & 76.99          & 85.21           \\
	\bottomrule
\end{tabular}
	\label{tab:ablation}
	\vspace{-1mm}
\end{table}

\begin{figure}[t]
	\centering
	\includegraphics[width=0.4\textwidth]{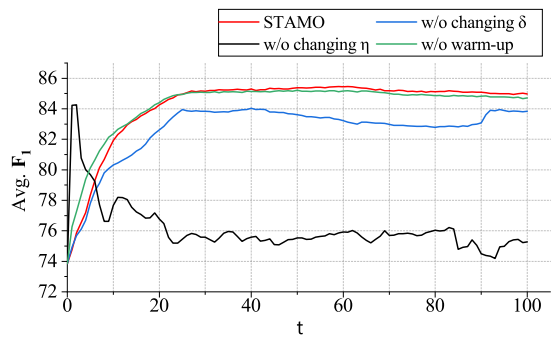}
		\vspace{-4mm}
	\caption{Average $\mbox{F}_1$-score as iterations progress and different modules are removed from inter-slot optimization separately.
	}
	\label{fig:ablation}
	\vspace{-3mm}
\end{figure}

In addition, it is observed that for the \textbf{STAMO} method, precision is the only metric decreasing in the iterative process. We believe it is caused by the error reinforcement problem. In our case, \enquote{errors} mainly means false positive predictions with respect to EEs. These wrong predictions would mislead the feature learning approach to regard wrong entities or words as relevant context of EEs, which further brings more false positive predictions. \textbf{STAMO} achieves the highest precision when convergence, which demonstrates its effectiveness in alleviating the error reinforcement problem.

\begin{table}[t]
%\begin{mdframed}[linecolor=blue,linewidth=1pt,innerrightmargin=6pt,innerbottommargin=6pt,innerleftmargin=6pt,innertopmargin=2pt,backgroundcolor=white]
		\caption{Ablation study of EL features}
		\label{tab:featablation}
		\centering
	\vspace{-3mm}
		\begin{tabular}{l|cccc}
			\toprule
			\multicolumn{1}{l}{\multirow{2}{*}{EL features}} & \multicolumn{4}{c}{{Avg.}}                                  \\
			\cmidrule(l){2-5}
			\multicolumn{1}{l}{}  & Acc          & $\Delta$Acc         & $\mbox{F}_1$          & $\Delta\mbox{F}_1$   \\
			\midrule
		{Three EL features} & \textbf{77.34}  & \textbf{-} & \textbf{85.45} & \textbf{-}\\
			w/o prior probability  & 50.26 & -27.08 & 56.50 & -28.95  \\ %\hline
			w/o relatedness score & 67.92 & -9.42 & 83.22 & -2.23 \\ %\hline
			w/o entity embdding &  63.44 & -13.90 & 72.37 & -13.08 \\ %\hline
			\bottomrule
		\end{tabular}
%	\end{mdframed}
 \vspace{-3mm}
%\vspace{-1.5mm}
\end{table}

\begin{figure}[t]
	\centering
	\includegraphics[width=0.45\textwidth]{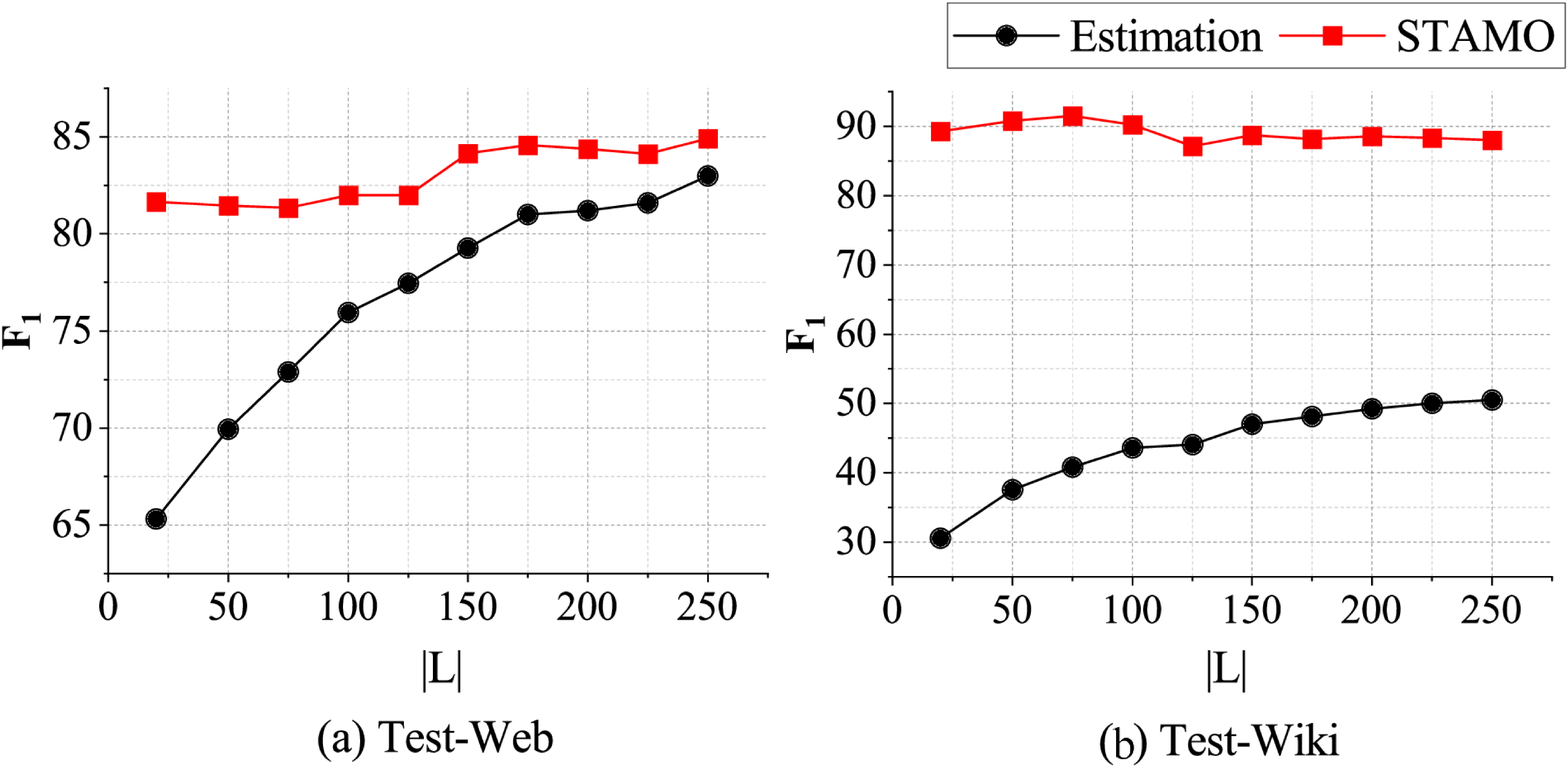}
	\vspace{-3mm}
	\caption{Effect analysis of the labeled data size.}
	\label{fig:effect_of_l}
	\vspace{-3mm}
\end{figure}

\subsection{Ablation Study}
\label{sec:ablation study}

We conduct an ablation study to verify the effectiveness of different modules applied in inter-slot optimization. There are three modules: (1) the adaptive adjustment of gradient $\delta$, whose original value is $\theta^{t-1} - \theta^t$; (2) setting learning rate $\eta$ to a small value, whose original value is 1; and (3) the warm-up scheduling. We remove each of them separately and the experimental results of \textbf{STAMO}+DeepED are shown in Table \ref{tab:ablation}. We can see that the adjustments of $\delta$ and $\eta$ are both important, while the warm-up step also makes a slight improvement.
We further investigate the change of average $\mbox{F}_1$-score on the two datasets as iterations progress. The experimental results {of \textbf{STAMO}+DeepED} are shown in Figure \ref{fig:ablation}. It can be seen that setting $\eta$ to a small value is a simple but effective way to keep the iterative process stable, which is similar to the effect of learning rate in the gradient descent method: a too large learning rate may never reach the optimal solution while a too small learning rate will take a long time to converge.
In addition, the adaptive adjustment of $\delta$ makes self-training converge to a better result. As analyzed by \cite{diederik2015adam}, the adaptive adjustment of $\delta$ can be regarded as the signal-to-noise ratio. Since noise represents uncertainty, we believe such an adaptive adjustment could help our approach focus on the confident part when updating feature parameters for EEs, and thus it could alleviate the error reinforcement problem implicitly.
\textcolor{blue}{Warm-up is usually used in batch gradient descent approaches to alleviate the problem of instable gradient at the beginning of training. While in our case, the problem of instable gradient may not be obvious and thus the contribution of warm-up seems slight from the experimental results.}

\textcolor{blue}{We also conduct an ablation study to show the effectiveness of different EL features (i.e., prior probability, relatedness score, and entity embedding) used in this paper. We remove each of them separately and the experimental results of \textbf{STAMO}+DeepED are shown in Table \ref{tab:featablation}. We can see that each of the three EL features has a positive contribution to the final performance and when all the three EL features are leveraged together, it yields the best performance, which is consistent with our intuition. It is also noted that our proposed \textbf{STAMO} could be flexibly integrated with any other numerical EL feature besides these three features.
}

\vspace{-2mm}

\subsection{Effect Analysis of Labeled Data Size}
\label{sec:effect analysis}

Finally, we aim at answering the following question: how does the number of labeled documents influence the quality of EEs' features? We vary the labeled data size $|L|$, and the experimental results {based on DeepED} are shown in Figure \ref{fig:effect_of_l}. It is observed that for the approach \textbf{Estimation}, the labeled data size has a significant impact on its results, while \textbf{STAMO} is much more robust. When reducing the labeled data size from 250 to 20, the $\mbox{F}_1$-score of \textbf{Estimation} changes $-17.66\%$ and $-19.98\%$ on \textbf{Test-Web} and \textbf{Test-Wiki} respectively, while the $\mbox{F}_1$-score of our proposed approach \textbf{STAMO} changes $-3.26\%$ and $+1.27\%$ respectively. It demonstrates that our approach is highly effective to leverage the knowledge hidden in the unlabeled data and it can greatly reduce the cost of human efforts.

%%%%%%%%%%%%%%%%%%%%%%%%%%%%%%%%%%%%%%%%
\vspace{-2mm}

\section{Related Work}

Two threads of research are related to our work, i.e., entity linking and self-training. We will introduce them in details.

\vspace{-2mm}

\subsection{Entity Linking}
Fully-supervised EL has been well studied by abundant literature \cite{shen2014entity,shen2021entity}, where \enquote{fully-supervised} means not only the training of EL models is supervised by some documents annotated for EL specifically (e.g., AIDA-CoNLL \cite{hoffart2011robust}), but also the estimation of EL features for entities is supervised by a labeled corpus (e.g., Wikipedia).
Specifically, prior probability provides the probability of the appearance of a candidate entity given the mention without considering the context where the mention appears. It is usually estimated using some structured information provided by Wikipedia such as hyperlinks in Wikipedia articles and pageview counts of the candidate entity. It plays a significant role in most EL methods \cite{ratinov2011local,hoffart2011robust,ganea2017deep,le2018improving,fang2020high, ran2018attention} due to its strong effectiveness.
Relatedness score (e.g., WLM \cite{witten2008effective}) is adopted by lots of EL works \cite{hoffart2011robust, ratinov2011local, shen2012linden,phan2018pair, xue2019neural},\textcolor{blue}{\cite{hoffart2012kore}} to calculate the topical coherence between candidate entities. WLM is defined under the assumption that two entities tend to be related if there are many Wikipedia articles that link to both.

Additionally, with the advent of deep learning, entity embedding that encodes each candidate entity as a dense and low-dimensional vector has been applied in many deep learning based EL methods \cite{shen2021entity,ganea2017deep,yamada2016joint,cao2017bridge, sevgili2019improving,fang2019joint,wu2020scalable,zwicklbauer2016robust} to implicitly represent the semantic and syntactic properties of the candidate entity. Rich data existing in Wikipedia (e.g., textual entity description \cite{ganea2017deep,fang2019joint,wu2020scalable}, entity context \cite{zwicklbauer2016robust,cao2017bridge}, and hyperlink structure of entity pages \cite{ganea2017deep,sevgili2019improving, yamada2016joint}) are usually leveraged to learn entity embeddings. To sum up, the estimation of these EL features for entities depends on various structures existing in Wikipedia.
% Recently, various methods based on random walk \cite{zwicklbauer2016robust, guo2018robust} and factor graph model \cite{ganea2016probabilistic, ran2018attention} have been studied. With the development of deep learning, neural network models including pairwise conditional random field (CRF) \cite{ganea2017deep, le2018improving, kolitsas2018end, chen2020improving}, graph neural networks \cite{cao2018neural, wu2020dynamic, fang2020high}, recurrent random walk networks \cite{xue2019neural}, and reinforcement learning \cite{fang2019joint} have also been investigated.
% On the other hand, entity embedding techniques also contribute to the improvement of EL. In practice, entity embeddings towards disambiguating mentions \cite{yamada-etal-2016-joint, cao2017bridge, ganea2017deep, cao2018joint} are different from similar techniques based on KB structures only \cite{bordes2013translating, lin2015learning}. They are usually trained with the objective of measuring relatedness between entities or predicting words appearing in contexts of entities.
% \cite{bordes2013translating, lin2015learning}
% based on previous works \cite{ganea2017deep,globerson2016collective}

Since manual annotation is time-consuming, weakly-supervised setting for EL was considered in recent works \cite{le2019boosting,le2019distant}. In this setting, KBs like Wikipedia are available while documents in the target domain are unlabeled (i.e., mentions are linked without any labeled examples). Specifically, Le and Titov \cite{le2019boosting} performed message passing based on the Wikipedia link subgraph corresponding to the document to construct a high recall list of candidates for each mention. What's more, Le and Titov \cite{le2019distant} leveraged entity type information provided by Freebase \cite{bollacker2008freebase} to generate entity embeddings for weakly-supervised linking.

% introduced the first approach to EL in an unsupervised setting, i.e., Wikipedia is not available and mentions are linked without any labeled examples. They framed EL as the multi-instance learning (MIL) problem and relied on surface matching to create initial noisy labels. They then designed a simple context similarity feature for linking and trained a binary classifier as the noise detection component to detect noisy data points since the surrogate labels are noisy.

In addition, zero-shot learning \cite{romera2015embarrassingly} has also been considered in the field of entity linking. The key idea of zero-shot is to train the model on a domain with rich labeled data resources and apply it to a new domain with minimal data \cite{sevgili2020neural}. In the setting of zero-shot EL \cite{li2020efficient,logeswaran2019zero,wu2020scalable}, the entity being linked with has not been seen during training and is only defined by a textual entity description from its corresponding Wikipedia page. Specifically, Logeswaran et al. \cite{logeswaran2019zero} and Wu et al. \cite{wu2020scalable} utilized a BERT-based \cite{devlin2019bert} cross-encoder to perform joint encoding of mentions and entities. The cross-encoder takes the concatenation of the mention context and the textual entity description to produce a scalar score for each candidate entity. Wu et al. \cite{wu2020scalable} and Li et al. \cite{li2020efficient} leveraged a BERT-based \cite{devlin2019bert} bi-encoder which uses two separate encoders to learn vectors of the mention context and the textual entity description respectively. The candidate entity ranking is performed by comparing dot-products of representation vectors.

In this paper, we consider a completely different setting from the settings introduced above. In our setting, EEs are newly discovered and have not be included in Wikipedia yet. Accordingly, no data from Wikipedia could be provided for EEs, which makes the existing EL models unable to link ambiguous mentions with EEs correctly as the absence of their EL features. Due to the fact that the world is constantly evolving and newly emerging entities are constantly being discovered, such a setting is essential and meaningful for real-world applications.
%  \cite{mintz2009distant}

\vspace{-2mm}
\subsection{Self-Training}
Self-training, or broadly speaking self-labeled, is an intuitive but successful methodology to tackle the semi-supervised classification problem. An overview of self-labeled techniques is presented in \cite{triguero2015self}. In general, traditional self-labeled methods aim at finding accurate confidence measures. For example, co-training \cite{blum1998combining} is a classic multi-view self-labeled technique. In co-training, each instance is represented by two sets of features, which are also called two views, and then two classifiers are separately trained on different views. If these two views are both redundant and conditionally independent, co-training can effectively reduce generalization errors. Recently, class-balanced self-training \cite{zou2018unsupervised} and confidence regularized self-training \cite{zou2019confidence} achieved state-of-the-art results in semantic segmentation.
Since instance selection based on the confidence score could be impractical in some scenarios, some recent works \cite{xie2020self, he2019revisiting} also began to focus on self-training without instance selection. In this paper, we creatively regard self-training as a multiple optimization process with respect to the EL features of EEs, and propose both intra-slot and inter-slot optimizations to alleviate the error reinforcement problem caused by the mislabeled data implicitly, rather than throw the mislabeled data away explicitly.

%%%%%%%%%%%%%%%%%%%%%%%%%%%%%%%%%%%%%%%%

\vspace{-2mm}
\section{Conclusions}

In this paper, we study a new task of learning EL features for emerging entities in a general way. We propose a novel approach \textbf{STAMO} to learn high-quality EL features for EEs automatically, which needs just a small number of labeled documents for each EE collected from the Web, as it could further leverage the knowledge hidden in the unlabeled data. \textbf{STAMO} is mainly based on self-training, which makes it flexibly integrated with any numerical EL feature or EL model. We regard self-training as a multiple optimization process and propose both intra-slot optimization and inter-slot optimization to alleviate the error reinforcement problem implicitly. In intra-slot optimization, we consider that the EL features of EEs should also minimize the objective function which is used to train the given EL model. In inter-slot optimization, we propose a hypothetical optimization process which makes a direct connection between the EL features of EEs in the current slot and the historical slots, and this novel perspective enables us to leverage the information provided by the historical slots to improve the future learning process. Finally, extensive experiments demonstrate the effectiveness of our approach.

\vspace{-2mm}
\section{Acknowledgements}
\label{ack}
\vspace{-1.5mm}
This work was supported in part by National Key Research and Development Program of China under Grant No. 2020YFA0804503, National Natural Science Foundation of China under Grant No. 61532010, and Beijing Academy of Artificial Intelligence (BAAI). Wei Shen was supported in part by National Natural Science Foundation of China (No. U1936206), YESS by CAST (No. 2019QNRC001), and CAAI-Huawei MindSpore Open Fund.

\vspace{-3mm}
%\bibliography{IEEEabrv,references}{}
%\bibliographystyle{IEEEtran}

\vspace{-10mm}

\begin{IEEEbiography}[{\includegraphics[width=1in,height=1.25in,clip,keepaspectratio]{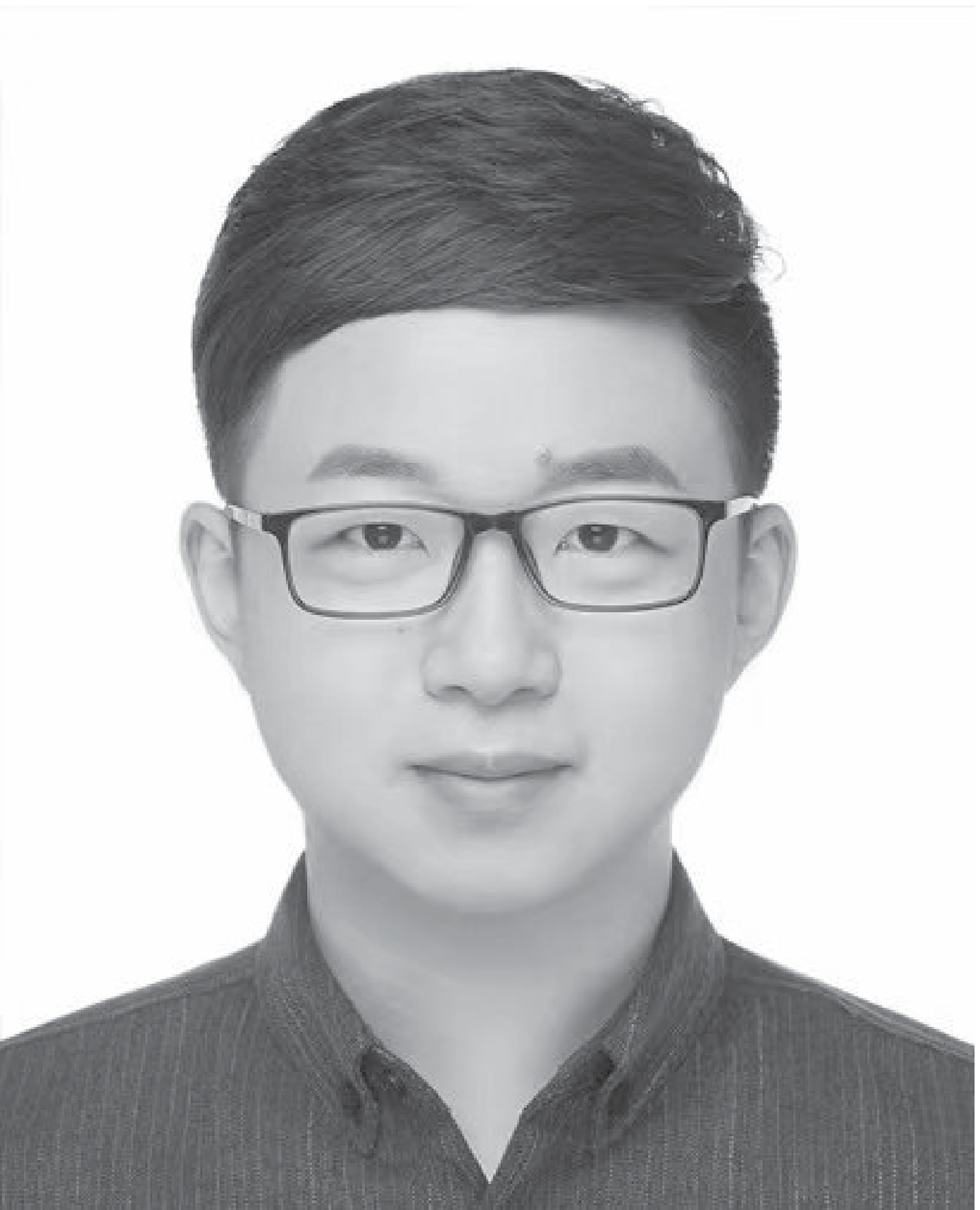}}]{Chenwei Ran} received the BS degree and the PhD degree in computer science from Tsinghua University, China, in 2014 and 2020 respectively. His research interests include entity linking, Web search, and natural language processing.

\end{IEEEbiography}
\vspace{-13.0mm}
% \vspace{-20mm}
\begin{IEEEbiography}[{\includegraphics[width=1in,height=1.25in,clip,keepaspectratio]{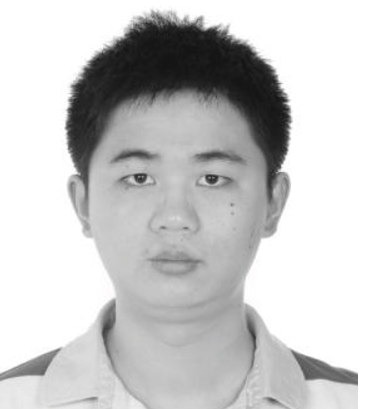}}]{Wei Shen} received the PhD degree in computer science from Tsinghua University, China, in 2014.
He is an associate professor in the College of Computer Science, Nankai University, China. His research interests include entity linking, knowledge base population, and text mining.
He was a recipient of ACM China Rising Star Award (Honorable Mention), CCF-Intel Young Faculty Researcher Program, and CAAI Outstanding Doctoral Dissertation Award.
\end{IEEEbiography}
\vspace{-13.0mm}
% \vspace{-100mm}

\begin{IEEEbiography}[{\includegraphics[width=1in,height=1.25in,clip,keepaspectratio]{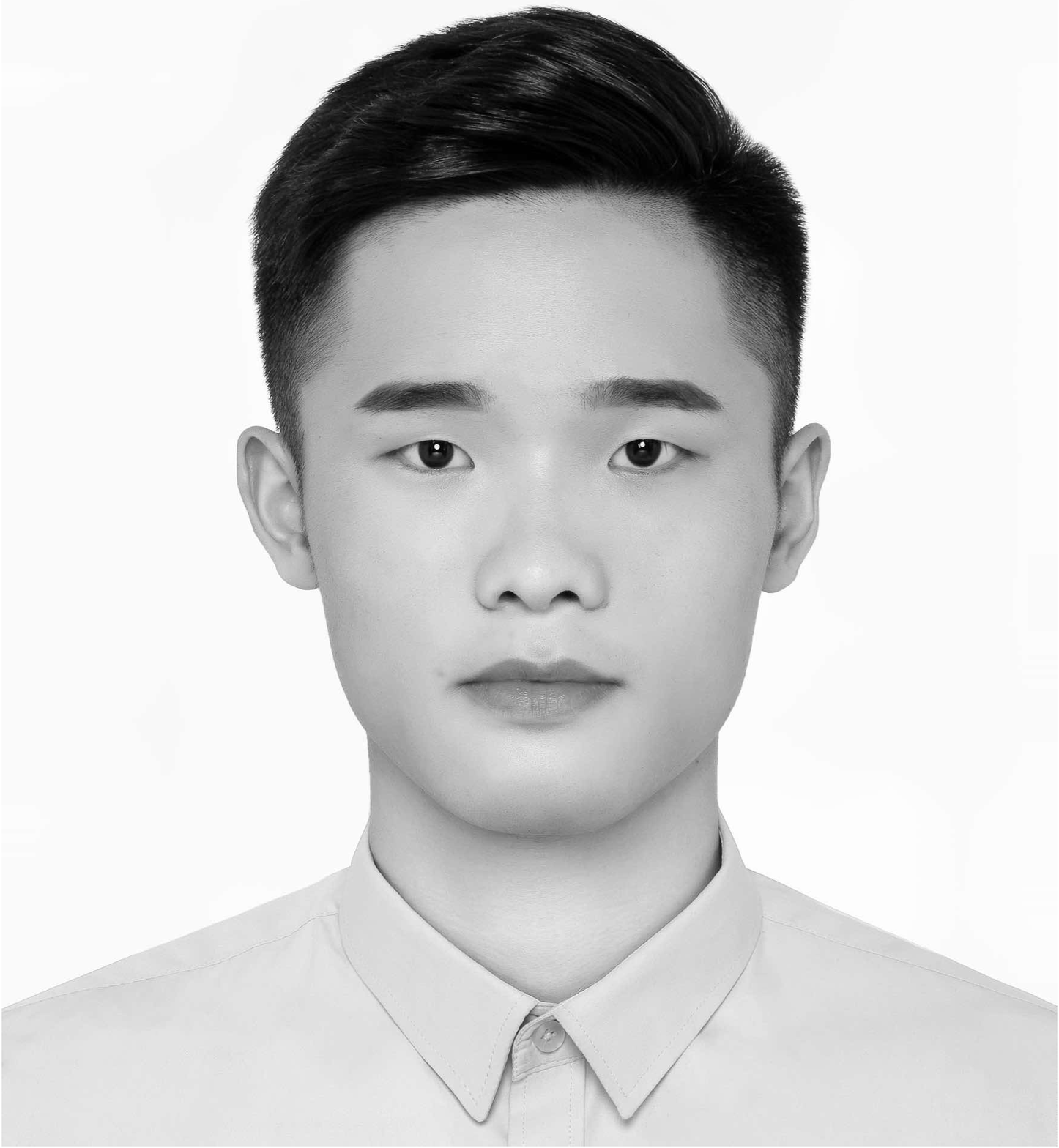}}]{Jianbo Gao} received his BS degree in computer science from Hefei University of Technology, China, in 2020. He is currently a master candidate at Nankai University. His research interests include data mining and knowledge graph.
	
\end{IEEEbiography}
\vspace{-13.0mm}

\begin{IEEEbiography}[{\includegraphics[width=1in,height=1.25in,clip,keepaspectratio]{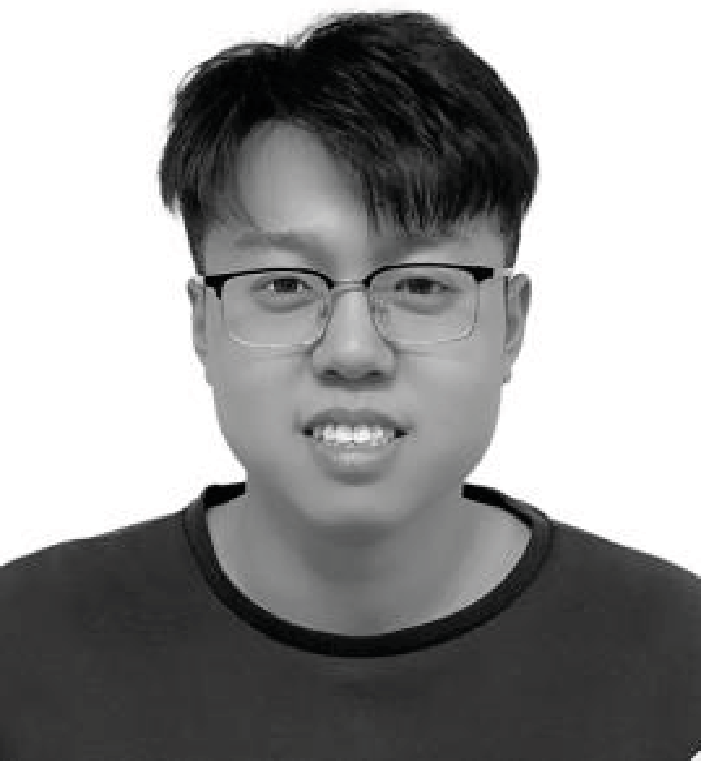}}]{Yuhan Li} received his BS degree from Northeast Forestry University, China in 2020. He is currently a master candidate at Nankai University. His research interests include knowledge graph, entity linking and data mining.

\end{IEEEbiography}
\vspace{-12.0mm}
% \vspace{-100mm}
\begin{IEEEbiography}[{\includegraphics[width=1in,height=1.25in,clip,keepaspectratio]{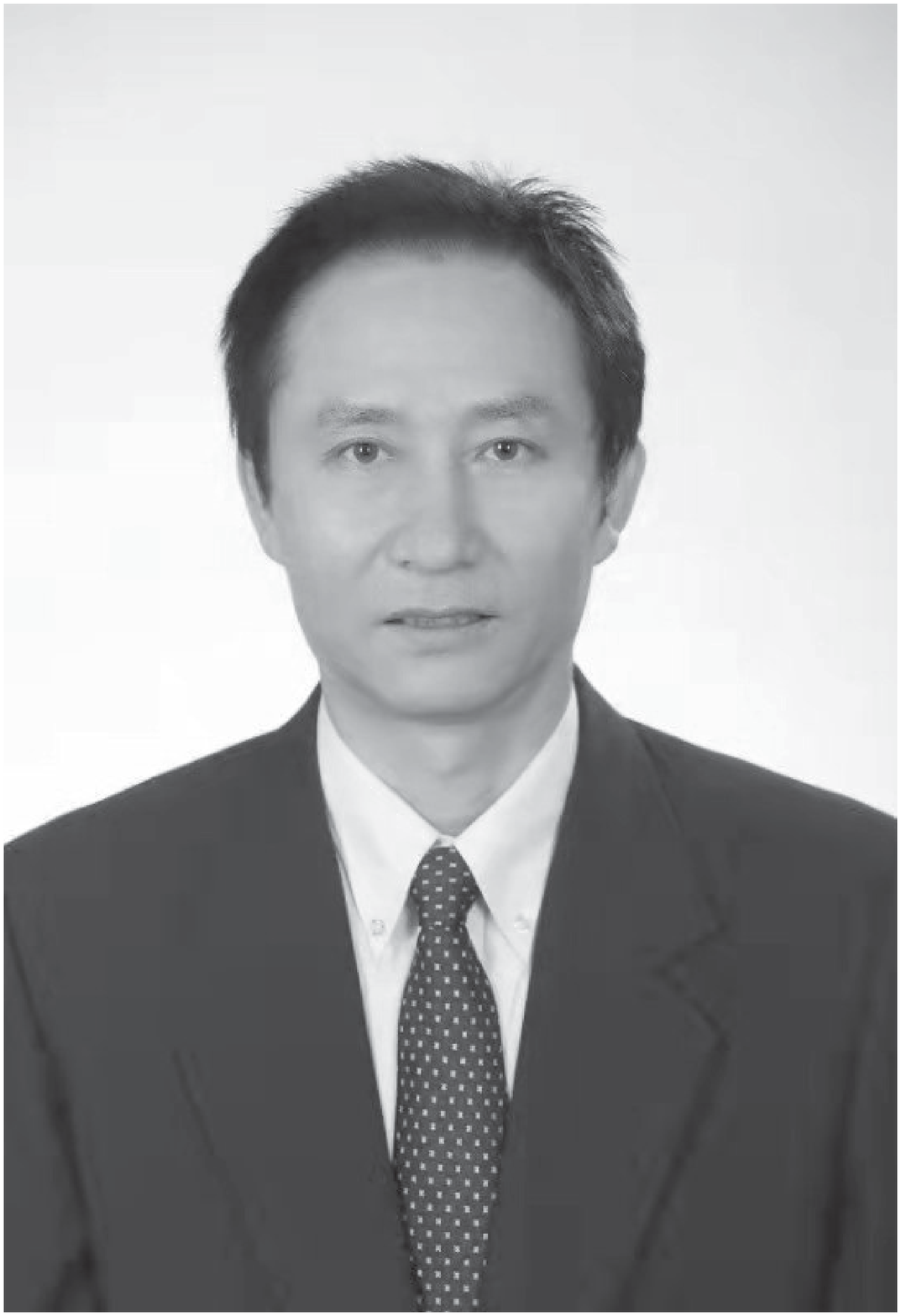}}]{Jianyong Wang}
 is currently a professor in the Department of Computer Science and Technology, Tsinghua University, Beijing, China.
 He received his PhD degree in Computer Science in 1999 from the Institute of Computing Technology, Chinese Academy of Sciences.
 His research interests mainly include data mining and Web information management. He has co-authored over 60 papers in some leading international conferences and some top international journals. He is serving or ever served as a PC member for some leading international conferences, such as SIGKDD, VLDB, ICDE, WWW, and an associate editor of IEEE TKDE and ACM TKDD. He is a Fellow of the IEEE, a member of the ACM.
\end{IEEEbiography}
\vspace{-12.0mm}
% \vspace{-100mm}
\begin{IEEEbiography}[{\includegraphics[width=1in,height=1.25in,clip,keepaspectratio]{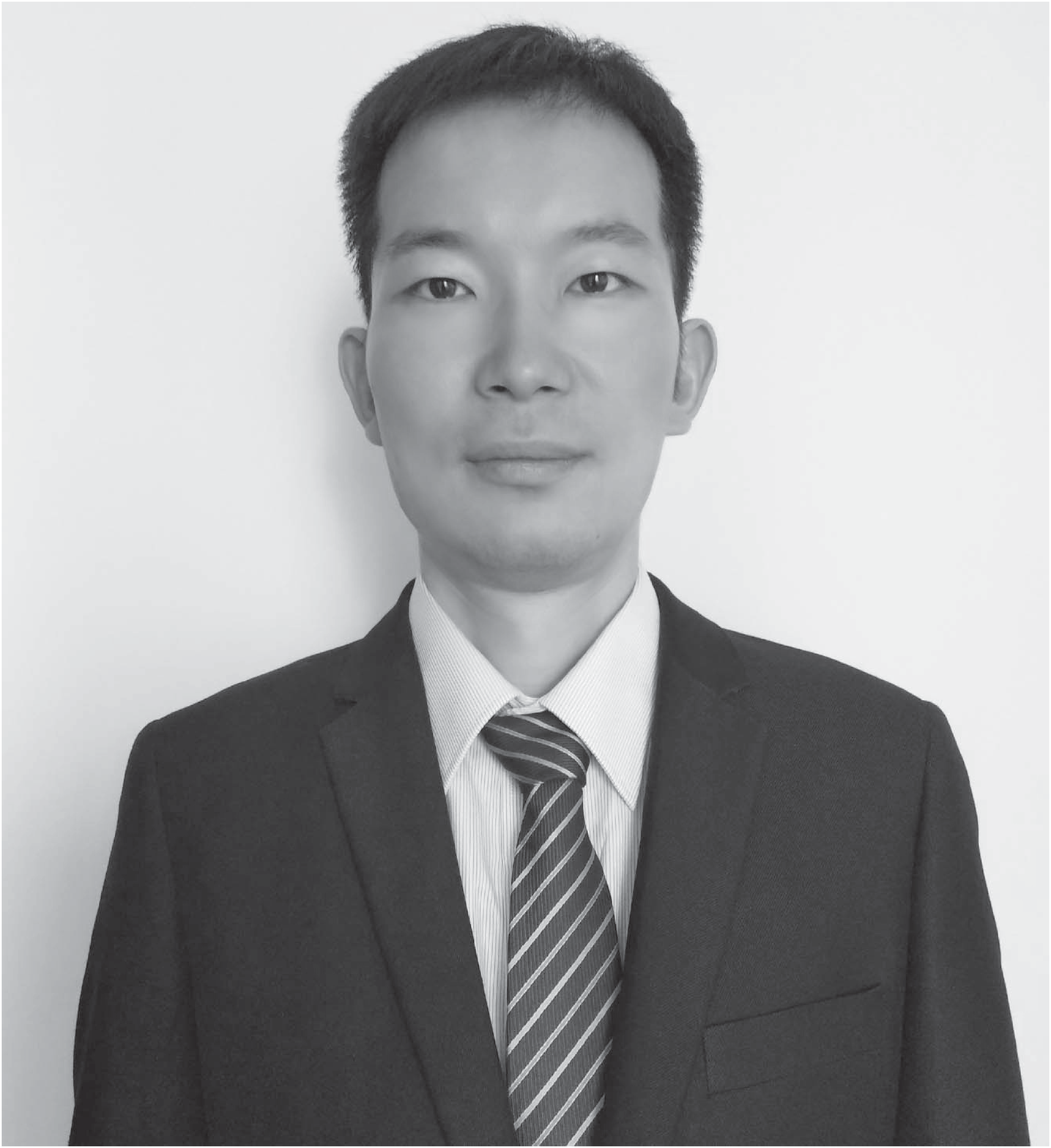}}]{Yantao Jia} is a technical expert in Huawei Technologies Co. Ltd. He received his PhD degree in mathematics. His main research interests include search, recommendation, open knowledge computing, social computing, and combinatorial algorithms.
\end{IEEEbiography}

\end{document}